\documentclass{talan-simple}

\usepackage[utf8]{inputenc} 
\usepackage{longtable}
\IfFileExists{algorithm.sty}{%
  \usepackage{algorithm}
}{%
  \newenvironment{algorithm}[1][]{%
    \begin{figure}[##1]%
  }{%
    \end{figure}%
  }%
}
\IfFileExists{algpseudocode.sty}{%
  \usepackage{algpseudocode}
}{%
  \newcounter{algline}
  \newenvironment{algorithmic}[1][]{%
    \begin{list}{\arabic{algline}.}{%
      \usecounter{algline}%
      \setlength{\leftmargin}{2.2em}%
      \setlength{\itemsep}{0.15em}%
    }%
  }{%
    \end{list}%
  }%
  \newcommand{\Require}{\item[\textbf{Require:}]}%
  \newcommand{\State}{\item}%
  \newcommand{\For}[1]{\item[\textbf{for} ##1 \textbf{do}]}%
  \newcommand{\EndFor}{\item[\textbf{end for}]}%
}

\title{\raisebox{-0.2\height}{%
\includegraphics[height=0.8em]{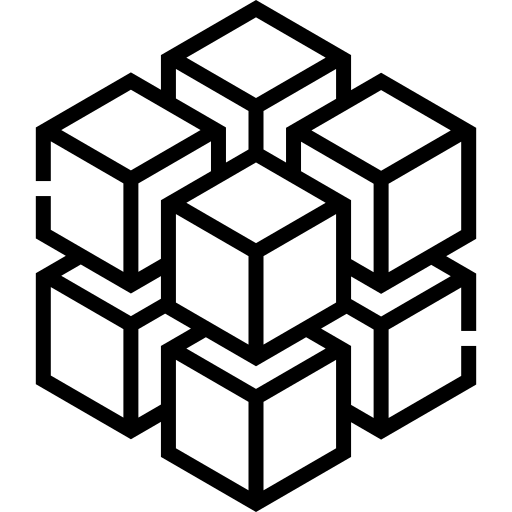}} {\fontsize{14pt}{20pt}\selectfont MC-RFM: Geometry-Aware Few-Shot Adaptation via Mixed-Curvature Riemannian Flow Matching} }
\author[*]{Salim Khazem}
\author[*]{Ibrahim Mohamed Serouis}
\author[*]{Zakaria Ezzahed}
\affiliation[*]{Talan Research Center, Paris, France}

\date{\today}
\correspondence{\email{salim.khazem@talan.com}, \email{ibrahim.mohamed-serouis@talan.com}, \email{zakaria.ezzahed@talan.com}}
\codeurl{\url{https://github.com/salimkhazem/MC-RFM}}

\definecolor{salimcolor}{RGB}{0,90,180}      
\definecolor{ibrahimcolor}{RGB}{180,80,0}    
\definecolor{zakariacolor}{RGB}{120,0,160}

\abstract{Parameter-efficient adaptation of pretrained vision models is commonly performed through linear probes, prompts, low-rank updates, or lightweight residual modules. While effective, these methods usually treat adaptation as a discrete Euclidean perturbation of frozen representations, without explicitly modeling the geometry of the task-induced feature displacement. We propose \textsc{MC-RFM}, a mixed-curvature Riemannian flow-matching framework for few-shot adaptation of frozen visual backbones. The key idea is to represent adapted features on a product manifold combining a hyperbolic factor, which captures hierarchy-sensitive semantic structure, and a Euclidean factor, which preserves locally discriminative visual variation. Adaptation is formulated as a task-conditioned continuous transport from frozen features to support-set prototypes, trained with a flow-matching objective and coupled to a hybrid prototype-linear classifier. The method is lightweight, backbone-agnostic, and operates entirely on cached frozen features. Across seven visual recognition benchmarks, five frozen backbones, and 1/4/16-shot regimes, \textsc{MC-RFM} is the best-performing method in a majority of evaluated settings, with the strongest gains on Transformer backbones and fine-grained datasets. Ablations show that the mixed-curvature head, task conditioning, adaptive branch gating, prototype shrinkage, and discriminative supervision each contribute to performance. These results suggest that few-shot adaptation benefits not only from deciding which parameters to update, but also from modeling how representations should move through a geometry matched to the structure of the downstream task.}

\begin{document}

\maketitle

\section{Introduction} \label{sec:introduction}
\begin{figure}[b]
    \centering
    \includegraphics[width=0.90\textwidth]{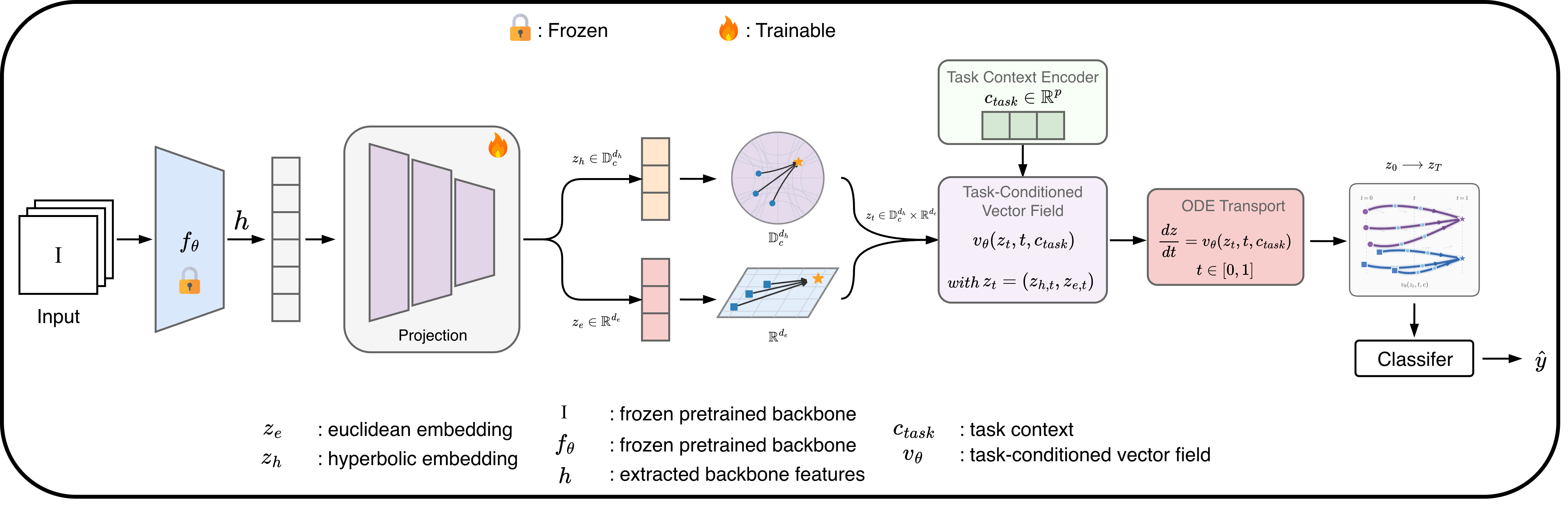}
    \caption{Overview of the proposed MC-RFM architecture. Given an input image \(I\), a frozen pretrained vision backbone \(f_\theta\) extracts a feature representation \(h\).
    A lightweight trainable projection module maps \(h\) into a mixed-curvature latent representation composed of a hyperbolic component \(z_h \in \mathbb{D}_c^{d_h}\) and a Euclidean component \(z_e \in \mathbb{R}^{d_e}\). A task context encoder produces \(c_{\mathrm{task}}\), which conditions the vector field \(v_\theta(z_t,t,c_{\mathrm{task}})\). The resulting ODE transport maps the initial state \(z_0\) to the transported state \(z_T\), which is then passed to the classifier to produce the prediction \(\hat{y}\). The lock icon denotes frozen modules, while the flame icon denotes trainable modules.}
    \label{fig:mcrfm_overview}
\end{figure}
Pretrained visual backbones \cite{he2016resnet,liu2022convnext,dosovitskiy2021vit,touvron2021deit,liu2021swin} have shifted few-shot adaptation toward \emph{re-using} frozen representations rather than learning them from scratch. Existing methods, including linear probing, prompt tuning, adapters, and low-rank updates \cite{houlsby2019adapters,hu2022lora,chen2022adaptformer}, mainly differ in \emph{which} parameters they update, but typically treat adaptation as a discrete Euclidean perturbation. This overlooks two aspects of downstream visual structure: visual classes may exhibit hierarchical relations that flat metrics distort~\cite{nickel2017poincare,khrulkov2020hyperbolic}, and adaptation is naturally viewed as a smooth transport from generic to task-specific features. Mixed-curvature representations address the former by combining hyperbolic and Euclidean factors \cite{gu2019mixedcurvature,skopek2020mixedvae,saezdeocarizborde2023nlgs}, while flow matching provides a simulation-free framework for learning transport vector fields \cite{lipman2023flowmatching,liu2023rectified}. Yet mixed-curvature methods are mostly static, and flow matching has been studied mainly for generative modeling. We combine these ideas in \textsc{MC-RFM}, a lightweight few-shot adapter that models adaptation as \emph{continuous, geometry-aware transport} of frozen features. MC-RFM projects features into \(\mathcal{M}=\mathbb{D}_c^{d_h}\times\mathbb{R}^{d_e}\), uses a task-conditioned vector field to transport them toward support-set prototypes through hyperbolic geodesic and Euclidean linear interpolation, and classifies the transported representation with an adaptively gated hybrid prototype-linear head. Training combines flow matching with cross-entropy supervision, and inference requires only a few ODE function evaluations.

\textbf{Contributions.} Our contributions are fourfold. 
(i) We introduce \textsc{MC-RFM}, a mixed-curvature Riemannian flow-matching framework that recasts few-shot adaptation of frozen visual backbones as task-conditioned continuous transport.  (ii) We design a feature-adaptive architecture that jointly modulates the hyperbolic--Euclidean representation balance, prototype--linear classifier balance, and transport dynamics while keeping hyperbolic states stable inside the Poincar\'e ball. 
(iii) We evaluate \textsc{MC-RFM} across seven benchmarks, multiple frozen backbones, and 1/4/16-shot regimes, showing strongest gains on fine-grained tasks with Transformer backbones.  (iv) Through ablations and stability diagnostics, we show that performance arises from the combination of mixed curvature, adaptive gating, task conditioning, prototype shrinkage, and joint flow-matching/classification supervision.


\section{Related work} \label{sec:related_work}

\textbf{Few-shot Adaptation of Pretrained Visual Models.}

Few-shot adaptation addresses learning from few labeled examples. Early approaches include metric- and optimization-based methods such as Matching Networks, Prototypical Networks, and MAML \cite{vinyals2016matching,snell2017prototypical,finn2017maml}. More recent work has shifted toward adapting strong pretrained visual backbones, including ResNet, ConvNeXt, ViT, DeiT, and Swin \cite{he2016resnet,liu2022convnext,dosovitskiy2021vit,touvron2021deit,liu2021swin}. Adaptation strategies range from linear probing and full fine-tuning to parameter-efficient methods such as adapters, LoRA, prompt tuning, and AdaptFormer and recent frozen-backbone adapter formulations \cite{houlsby2019adapters,hu2022lora,khazem2026topolora,chen2022adaptformer,khazem2026adaptertune}. While these methods differ in efficiency and robustness, they remain largely parameter-centric: they specify which parameters to update, but do not explicitly model the geometry or dynamics of feature adaptation.

\textbf{Euclidean, Hyperbolic, and Mixed-Curvature Representation Learning.}
Euclidean spaces remain standard in visual learning because they capture local appearance variation, smooth interpolation, and near-linear decision boundaries \cite{chen2020simclr,he2020moco,khosla2020supcon,radford2021clip}. However, visual categories often contain taxonomic, coarse-to-fine, or part-whole structure that flat metrics can distort \cite{nickel2017poincare,nickel2018lorentz}. Hyperbolic geometry addresses this through exponential volume growth and has improved classification, retrieval, zero-shot recognition, dense prediction, metric learning, and vision-language representations \cite{ganea2018hyperbolicnn,chami2019hgcn,liu2019hgnn,khrulkov2020hyperbolic,liu2020hyperbolicvisual,atigh2022hyperbolicsegmentation,ermolov2022hyperbolicvit}. Yet visual adaptation is not purely hierarchical: few-shot features also encode texture, pose, and intra-class appearance variation. Mixed-curvature product manifolds therefore combine hyperbolic and Euclidean factors to capture global semantic hierarchy and local visual variation jointly \cite{gu2019mixedcurvature,skopek2020mixedvae}, consistent with data-dependent latent geometry \cite{saezdeocarizborde2023nlgs}. Existing mixed-curvature methods remain largely static and do not model how frozen features should move toward task-specific few-shot targets, motivating our dynamic, geometry-aware transport formulation.


\textbf{Flow Matching and Feature-Space Transport.} 
Flow matching \cite{lipman2023flowmatching} trains continuous normalizing flows without simulation by regressing time-dependent vector fields along prescribed probability paths. Rectified and conditional variants improve this framework through straighter or conditional paths \cite{liu2023rectified,tong2024cfm}, while Riemannian flow matching extends it to manifolds using tangent- or chart-coordinate velocities \cite{chen2024riemannian}. However, these methods primarily target data-space generative modeling. Related ODE- and score-based feature-refinement methods remain Euclidean and do not exploit hierarchical class structure \cite{song2021scorebased,karras2022elucidating}. \textsc{MC-RFM} instead applies Riemannian flow matching on a mixed-curvature product manifold, using learned transport as a task-conditioned feature-space adapter for frozen visual backbones, so adaptation becomes continuous geometric transport rather than a discrete PEFT-style update.

\section{Proposed method : MC-RFM} \label{sec:proposed_method}
\begin{figure}[h]
    \centering
        \centering
        \includegraphics[width=0.60\textwidth]{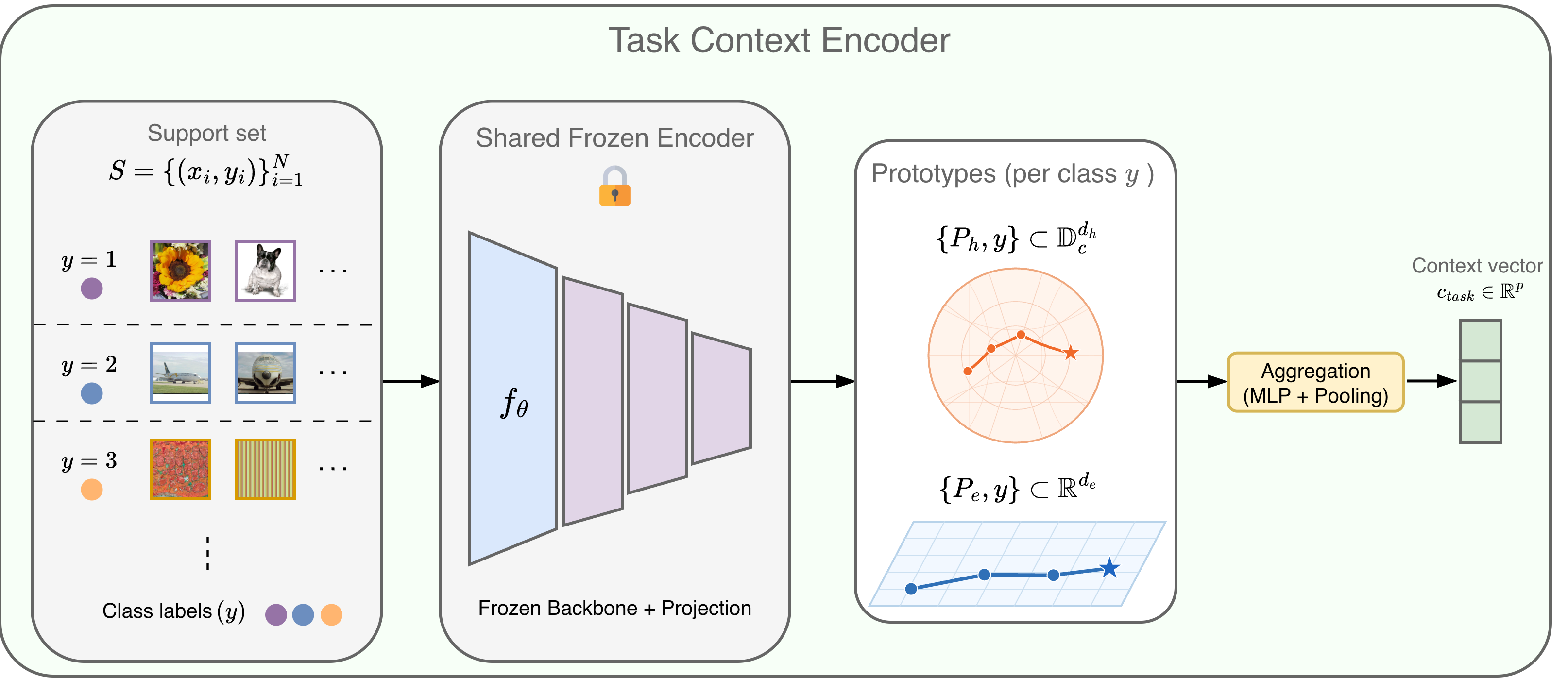}
        \caption{Detailed task context encoder. The support set \(S=\{(x_i,y_i)\}_{i=1}^{N}\) is processed by the shared frozen backbone and projection module. Class-wise prototypes are estimated in both the hyperbolic branch 
        \(\{p_{h,y}\}\subset\mathbb{D}_c^{d_h}\) and the Euclidean branch \(\{p_{e,y}\}\subset\mathbb{R}^{d_e}\).
        These prototypes are aggregated through a lightweight pooling module to produce the task context vector \(c_{\mathrm{task}}\in\mathbb{R}^{p}\).}
        \label{fig:task_context_encoder}

\end{figure}
\begin{figure}[h]
        \centering
        \includegraphics[width=0.60\textwidth]{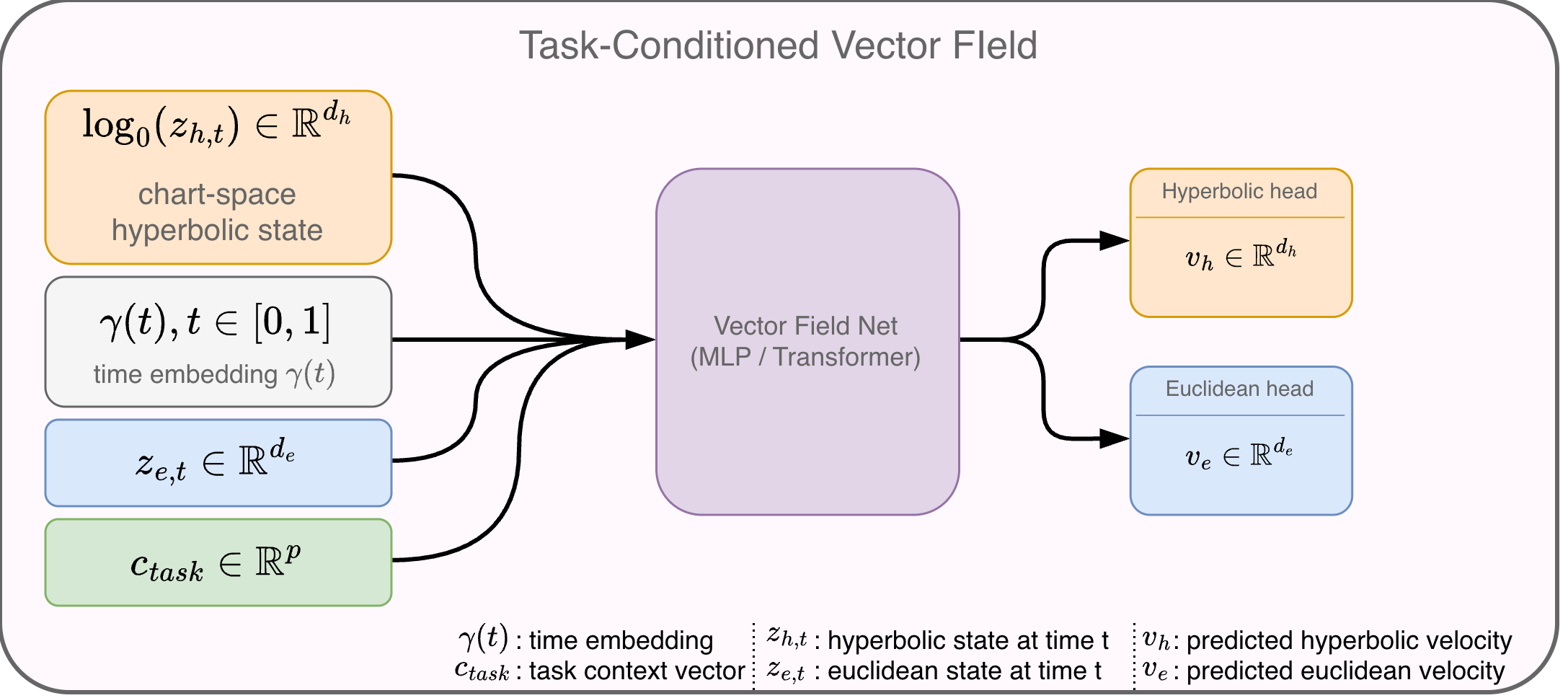}
        \caption{Detailed task-conditioned vector field. The vector field receives the chart-space hyperbolic state \(\log_0(z_{h,t})\), the Euclidean state \(z_{e,t}\), the time embedding \(\gamma(t)\), and the task context vector \(c_{\mathrm{task}}\). A shared vector-field network predicts branch-specific velocities
        \(v_h\) and \(v_e\), corresponding respectively to the hyperbolic and Euclidean components of the product manifold.}
        \label{fig:task_conditioned_vector_field}
\end{figure}

\subsection{Problem statement}
\label{sec:problem}

We consider few-shot adaptation of a frozen visual backbone. Let \(f_{\psi}: \mathcal{X} \rightarrow \mathbb{R}^{d}\) denote a pretrained feature extractor with fixed parameters \(\psi\). For a downstream task with a support set \(\mathcal{S} = \{(x_{i},y_{i})\}_{i=1}^{n}, \qquad y_{i} \in \{1, \ldots,K\}\), and query set \(\mathcal{{Q}}\), the goal is to learn a lightweight task-specific adapter on top of cached features \(h_{i}=f_{\psi(x_{i})}\), without updating the backbone. 

Standard linear probing learns a decision boundary directly in \(\mathbb{R}^{d}\). In contrast, we learn a continuous transport map that moves frozen features toward class-conditioned targets. The central hypothesis is that few-shot visual adaptation benefits from separating two kinds of structure: (i) a hierarchy-sensitive component and (ii) a locally discriminative component. We therefore define the adapter on the product manifold: 
\begin{equation}
\mathcal{M} = \mathbb{D}_{c}^{d_{h}} \times \mathbb{R}^{d_{e}}, \qquad d_{h}+d_{e}=m   
\end{equation}

where \(\mathbb{D}_{c}^{d_{h}}\) is the Poincare ball with curvature \(-c\), and \(\mathbb{R}^{d_{e}}\) is a Euclidean factor. The hyperbolic branch is intended to represent a hierarchy-like class organization, while the Euclidean branch captures residual local variation. 

\subsection{Mixed-Curvature Feature Parameterization}
\label{subsec:mixed_parameterization}
Given a frozen feature \(h=f_\psi(x)\), MC-RFM first applies a lightweight bottleneck projection into two branches:
\(u_h^{\mathrm{raw}} = W_{h} h + b_h, \qquad 
u_e^{\mathrm{raw}} = W_{e} h + b_e\). The hyperbolic branch is normalized and scaled before being mapped to the Poincare ball:
\[
\bar{u}_h = \frac{u_h^{\mathrm{raw}}}{\|u_h^{\mathrm{raw}}\|_2+\varepsilon},
\qquad
u_h = \alpha_h \bar{u}_h,
\qquad
z_h = \exp_0^c(u_h).
\]
The scalar \(\alpha_h\) is learned but constrained to a safe interval \(\alpha_h \in [\alpha_{\min}, \alpha_{\max}]\), which prevents the initialization from placing points close to the Poincare boundary. The Euclidean branch uses independent normalization \(u_e = \alpha_e \operatorname{LN}(u_e^{\mathrm{raw}}),
\ \
z_e = u_e\), where \(\alpha_e>0\) is learned. The resulting latent state is \(z = (z_h,z_e) \in \mathbb{D}_{c}^{d_h} \times \mathbb{R}^{d_e}\).

This parameterization is deliberately conservative. The hyperbolic branch is given enough capacity to encode negative-curvature structure, but its norm is controlled so that optimization begins in a well-conditioned region of the ball.

\subsection{Class Prototypes and Task Context}
\label{subsec:prototypes_task_context} 
For each task, class targets are constructed from the support set. We compute support embeddings with the current adapter projector and form bottleneck-space prototypes:
\[
\mu_{h,k} = \frac{1}{|\mathcal{S}_k|}
\sum_{(x_i,y_i)\in \mathcal{S}_k} u_h(x_i),
\qquad
\mu_{e,k} = \frac{1}{|\mathcal{S}_k|}
\sum_{(x_i,y_i)\in \mathcal{S}_k} u_e(x_i),
\]
where \(\mathcal{S}_k=\{(x_i,y_i)\in \mathcal{S}:y_i=k\}\). To reduce low-shot variance, we shrink class prototypes toward the global support prototype \(\tilde{\mu}_{h,k} = (1-\tau)\mu_{h,k}+\tau \bar{\mu}_h,
\qquad
\tilde{\mu}_{e,k} = (1-\tau)\mu_{e,k}+\tau \bar{\mu}_e,\) with shrinkage coefficient \(\tau \in [0,1]\). The product-manifold prototype for class \(k\) is:
\begin{equation}
p_k = (p_{h,k},p_{e,k})
=\left(\exp_0^c(\tilde{\mu}_{h,k}),\tilde{\mu}_{e,k}\right)    
\end{equation}

The same prototype bank is also used to condition the transport dynamics. We first map hyperbolic prototypes back to the origin chart:\(\xi_{h,k}=\log_0^c(p_{h,k}),
\ \
\xi_k = [\xi_{h,k};p_{e,k}]\). A task encoder summarizes the prototype set using token-wise projection, attention pooling, and global statistics. In particular, the context includes branch norms, pairwise prototype distances, and the number of classes. The output is a compact vector \(c_{\mathcal{S}} = E_{\eta}(\{p_k\}_{k=1}^{K}) \in \mathbb{R}^{d_c}\). This context makes the vector field task-conditioned rather than purely task-agnostic.

\subsection{Task-Conditioned Flow Matching}
\label{subsec:flow_matching}

For a labeled support example \((x_{i},y_{i})\), let  \(z_0^i=(z_{h,0}^i,z_{e,0}^i)\)
be its initial mixed-curvature representation and let \(z_1^i=p_{y_i}\) be the prototype of its class. We define a product path between source and target:
\(z_t^i = \gamma_{\mathcal{M}}(z_0^i,z_1^i;t)
=
\left(
\gamma_{\mathbb{D}_c}(z_{h,0}^i,z_{h,1}^i;t),
(1-t)z_{e,0}^i+t z_{e,1}^i
\right),
\ \ t\sim \mathcal{U}(\varepsilon,1-\varepsilon)\). The hyperbolic path uses Poincare geodesic interpolation, while the Euclidean path is linear. The vector field is parameterized as \(v_\theta(z_t,t,c_{\mathcal{S}})
=
\left(v_{\theta,h}(z_t,t,c_{\mathcal{S}}),v_{\theta,e}(z_t,t,c_{\mathcal{S}})\right)\). In practice, the hyperbolic input to the network is expressed in the origin chart \(\chi_h(t)=\log_0^c(z_{h,t})\), and the network receives \([\chi_h(t);z_{e,t};\phi(t);c_{\mathcal{S}}]\), where \(\phi(t)\) is a sinusoidal time embedding. This chart-space input improves conditioning while the ODE state itself remains on the product manifold.

The default target velocity is also computed in the stable origin chart:
\[
u_h^\star(t)
=
\frac{\log_0^c(z_{h,1})-\log_0^c(z_{h,t})}{1-t},
\qquad
u_e^\star(t)
=
\frac{z_{e,1}-z_{e,t}}{1-t}.
\]
The flow-matching loss is
\begin{equation}
\begin{split}
\mathcal{L}_{\mathrm{FM}}
=\,&
\mathbb{E}_{(x,y),t}
\Big[
w_h(t)\,m_h(z_t,c_{\mathcal{S}})
\left\|v_{\theta,h}(z_t,t,c_{\mathcal{S}})-u_h^\star(t)\right\|_2^2 \\
&\quad
+\,\lambda_e\,m_e(z_t,c_{\mathcal{S}})
\left\|v_{\theta,e}(z_t,t,c_{\mathcal{S}})-u_e^\star(t)\right\|_2^2
\Big].
\end{split}
\end{equation}
where \(w_h(t)\) is an epoch-dependent warmup/ramp schedule for the hyperbolic branch, and \(m_h,m_e\) are adaptive branch multipliers described below.

\subsection{Adaptive Branch Gating}
\label{subsec:adaptive_gating}

MC-RFM uses an adaptive gate to control how much each sample relies on the hyperbolic and Euclidean factors. For a transported state \(z=(z_{h},z_{e})\), we form normalized branch features \(r_{h} = \operatorname{LN}(\log_0^c(z_h))\), 
\(r_e = \operatorname{LN}(z_e)\), and compute a gate \(g(z,c_{\mathcal{S}})=
\sigma\left(
\rho_g + \Delta_g([r_h;r_e]) + b_g(c_{\mathcal{S}})
\right)
\in [0,1]\). The gate is converted into branch multipliers \(m_h = 2g\) and \(m_e = 2(1-g)\). These multipliers are used in two places. First, they weight the hyperbolic and Euclidean flow-matching terms. Second, they scale branch contributions in the classifier. This means the model can reduce the influence of an unhelpful branch for a given task or sample without removing that branch globally. This is a contribution-level detail but also a limitation: the current gate modulates branch contribution in the loss and classifier, but it does not yet route through different ODE architectures.

\subsection{Hybrid Prototype-Linear Classifier}
\label{subsec:hybrid_classifier}
After transport, MC-RFM predicts labels with a hybrid head. The prototype component uses calibrated product distances 
\(\ell_k^{\mathrm{proto}}(z)
=
-
\left(
m_h\,\gamma_h\,d_{\mathbb{D}_c}(z_h,p_{h,k})^2
+
m_e\,\gamma_e\,\|z_e-p_{e,k}\|_2^2
\right)\),where \(\gamma_h=\operatorname{softplus}(\rho_h)\) and \(\gamma_e=\operatorname{softplus}(\rho_e)\) are learned positive calibration parameters. The linear component operates on the chart-space transported representation:

\begin{equation}
\ell^{\mathrm{lin}}(z)=W_c\,[\sqrt{m_h}\,\operatorname{LN}(\log_0^c(z_h));
\sqrt{m_e}\,\operatorname{LN}(z_e)]+b_c    
\end{equation}

The final logits are:
\begin{equation}
\ell(z)=\beta(z,c_{\mathcal{S}})\ell^{\mathrm{proto}}(z)
+
\left(1-\beta(z,c_{\mathcal{S}})\right)\ell^{\mathrm{lin}}(z)   
\end{equation}

where \(\beta(z,c_{\mathcal{S}})
=
\sigma\left(\rho_\beta+\Delta_\beta([r_h;r_e])+b_\beta(c_{\mathcal{S}})\right)\). Thus, the model can interpolate between metric-based prototype inference and a discriminative linear head.

\subsection{Training Objective}
\label{subsec:training_objective}

The full training objective combines flow matching and discriminative supervision:
\begin{equation}
\mathcal{L}
=
\mathcal{L}_{\mathrm{FM}}
+
\lambda_{\mathrm{cls}}\,
\mathcal{L}_{\mathrm{CE}}(\ell(z_T),y)    
\end{equation}

where \(z_T\) is obtained by integrating the learned vector field from \(t=0\) to \(t=1\). The cross-entropy term uses label smoothing. This auxiliary supervision is important in the few-shot setting because pure prototype transport can be brittle when support prototypes are noisy. At inference time, the support prototypes and task context are recomputed from the support set. Query features are projected to \(\mathcal{M}\), transported with a low-NFE fixed-step ODE solver, and classified with the hybrid head.

\subsection{Theoretical Properties and Scope}
\label{subsec:theory}

\paragraph{Proposition 1: Interior stability of the hyperbolic state.}
Assume \(c>0\) and that every hyperbolic update is followed by the projection
\[
\Pi_{\varepsilon}(z) =
\begin{cases}
z, & \sqrt{c}\|z\|_2 \leq 1-\varepsilon,\\
\frac{1-\varepsilon}{\sqrt{c}}\frac{z}{\|z\|_2}, & \text{otherwise}.
\end{cases}
\]
Then all hyperbolic states produced by the discrete solver satisfy \(\sqrt{c}\|z_h\|_2 \leq 1-\varepsilon\).
Consequently, \(\exp_0^c\), \(\log_0^c\), Poincare distances, and conformal factors are evaluated away from the singular boundary.

\textbf{Proof sketch.}
The statement follows directly by induction over solver steps. The initial state is produced by \(\exp_0^c\) and projected to the ball. If the state at step \(s\) is finite, the solver produces a finite candidate update. Applying \(\Pi_\varepsilon\) maps this candidate into the closed interior ball of radius \((1-\varepsilon)/\sqrt{c}\). Therefore the invariant holds at step \(s+1\).

\textbf{Proposition 2: Flow-matching target consistency in the origin chart.}
For the Euclidean branch and the origin-chart hyperbolic branch, define \(\xi_h(t)=\log_0^c(z_{h,t}),\ \ \xi_e(t)=z_{e,t}\). The target velocity used by MC-RFM is the constant-speed velocity of the straight path from \((\xi_h(t),\xi_e(t))\) to \((\xi_h(1),\xi_e(1))\) in chart coordinates:
\begin{equation}
u^\star(t)
=
\frac{(\xi_h(1),\xi_e(1))-(\xi_h(t),\xi_e(t))}{1-t} 
\end{equation}

Thus, minimizing the population flow-matching loss recovers the conditional mean target velocity within the chosen chart-space parameterization.

\textbf{Proof sketch.} For squared-error regression, the population minimizer is the conditional expectation of the target velocity given the model inputs. Since MC-RFM trains \(v_\theta\) by squared error against \(u^\star(t)\), an expressive vector field recovers this conditional velocity in the origin chart. This is the standard regression property of flow matching; the approximation is that the hyperbolic branch uses a stable origin chart rather than an exact tangent field at every \(z_{h,t}\).

\section{Experiments} \label{sec:experiments}

\subsection{Experimental setup}
All experiments were conducted on a shared compute infrastructure comprising one NVIDIA RTX 6000 Ada Generation GPU with 49\,GB of VRAM and two NVIDIA RTX 5000 GPUs with 32\,GB of VRAM each. The system possessed an Intel Xeon w5-3435X (4.70\,Ghz) CPU featuring 32 threads. The complete training and inference procedures are provided in Appendix~\ref{app:algorithms}.

\paragraph{Backbones and shots}
We adapt five publicly available frozen backbones spanning convolutional and Transformer families: ResNet-50 \cite{he2016resnet}, ConvNeXt-Tiny and ConvNeXt-Base \cite{liu2022convnext}, ViT-B/16 \cite{dosovitskiy2021vit}, DeiT-Base \cite{touvron2021deit}, and Swin-Tiny \cite{liu2021swin}. For each backbone, we evaluate three few-shot regimes ($K\in\{1,4,16\}$ shots per class), giving up to 18 cells per dataset.

\paragraph{Baselines}
We compare \textsc{MC-RFM} against two controlled variants sharing the same projector, classifier head, training schedule, and seeds. \textsc{Euclidean} removes the hyperbolic factor by operating entirely in $\mathbb{R}^{d}$, while \textsc{Hyperbolic-only} removes the Euclidean branch and operates entirely on the Poincar\'e ball. These baselines isolate the effect of the hyperbolic prior and the benefit of mixed hyperbolic--Euclidean geometry.

\paragraph{Reproducibility and experiments protocol}
We use a fixed protocol across all datasets, backbones, shot counts, and methods. Few-shot support indices are sampled once per seed and stored to disk, so all methods see identical splits. Results are reported as the mean and standard deviation over three seeds $\{42,43,44\}$, controlling both support sampling and adapter initialization, with deterministic PyTorch behavior enabled. Backbones are frozen, and features are cached on disk. All models are trained for 50 epochs with AdamW (lr $5{\times}10^{-4}$, weight decay $10^{-4}$, batch-size 64, 5 warm-up epochs, and cosine decay; gradient clipping 1.0); flow-based methods use an Euler solver with 3 function evaluations.

\paragraph{Datasets}

\begin{sloppypar}
We evaluate our approach on fine-grained benchmarks from the literature such as FGVC-Aicraft \cite{maji_fine-grained_2013} and Flowers102 \cite{nilsback_automated_2008}, standard object recognition (CIFAR-10 and CIFAR-100 \cite{krizhevsky_learning_2009}), texture-focused classification with DTD \cite{cimpoi_describing_2014}, aerial scene understanding with EuroSAT \cite{helber_eurosat_2019}, and large-scale food recognition on Food-101 \cite{bossard_food-101_2014}.  These datasets covering varying levels of intra-class variation, inter-class similarity, semantic granularity, and latent hierarchy structure, provide us a comprehensive testbed for evaluating the robustness and generalization ability of our approach.   
\end{sloppypar}

\subsection{Main results}

Results in Table~\ref{tab:main_4shot_results} indicate that MC-RFM outperforms competing approaches in most experimental settings.  More global results in Appendix \ref{app:complete_results} confirm this statement. The largest improvement reaches approximately $2\%$ accuracy over the second-best method, with an average gain of roughly $1\%$ across all comparisons, even on harder tasks. However, these aggregate numbers mask substantial heterogeneity: the effect of MC-RFM varies differently across backbones and dataset tasks. To unravel these factors, we stratify the results along \textbf{(1) Architectures} where we distinguish \emph{Transformer-based} backbones (ViT, DeiT, Swin) from \emph{convolutional} backbones (ConvNeXt, ResNet-50), and \textbf{(2) Task type} where we partition the seven datasets into three mutually exclusive groups: 

\textbf{Coarse-grained, low-to-moderate latent hierarchy.} These tasks have relatively low intra-class variability and are mainly solved through global structure rather than fine local cues, as overall shape and layout dominate texture. This group includes \textbf{CIFAR-10} and \textbf{CIFAR-100}.

\textbf{Fine-grained, high hierarchy.} These tasks contain visually similar categories with subtle inter-class differences, frequent class overlap, and higher intra-class variation, making local discriminative cues particularly important. This group includes \textbf{FGVC-Aircraft} (aircraft model distinctions), \textbf{Flowers102} (high overlap between flower species), and \textbf{Food101} (strong variation due to lighting setups and presentation).

\textbf{Scene/texture-oriented.} These tasks emphasize local texture statistics or scene materials more than object identity. This group includes \textbf{DTD} (abstract texture categorization) and \textbf{EuroSAT} (remote-sensing scene labels sometimes texture-dominant, e.g., farmland, forest).


In the following sub-sections, we analyze performance across architectures, task types, and shot counts. A \textbf{positive contribution} denotes cases where \textsc{MC-RFM} performs best, measured by its margin over the strongest baseline, i.e., the second-best method. A \textbf{negative contribution} denotes cases where \textsc{MC-RFM} is not best, measured by its gap to the best-performing method.

\subsubsection{Focus on model and task type}

When stratifying results by architecture and task type, clear trends emerge. Transformer-based backbones exhibit a strong affinity with \textsc{MC-RFM}: in $75\%$ of settings they surpass the strongest alternative, whereas convolutional backbones do so in only $45\%$ of cases and display an overall negative net contribution, with the largest drop reaching $-3.07\%$. This effect is further amplified in the fine-grained, high-hierarchy task type. For transformer backbones, \textsc{MC-RFM} yields positive contributions in $83.33\%$ of fine-grained experiments (20/24), substantially higher than in the scene/texture-oriented ($72\%$) and coarse-grained ($66.66\%$) task types. A qualitatively similar result holds for convolutional backbones, albeit with weaker gains. However, the above analysis aggregates across shot counts. This raises a key question: does \textsc{MC-RFM} benefit systematically from increasing the number of shots, and does this interaction depend on architecture family and task type?

\subsubsection{Impact of shots}
Shot count reveals a strong architecture-dependent effect. Overall, \textsc{MC-RFM} remains competitive in the 1-shot regime, with positive contributions in $64.28\%$ of settings. For Transformer backbones, it is consistently effective, reaching $80.95\%$ positive contributions at 1 shot and $77.77\%$ at 16 shots, with the strongest behavior on fine-grained 16-shot tasks, where it is best in all cases ($100\%$). Convolutional backbones behave differently: \textsc{MC-RFM} yields a net positive effect only at 16 shots ($61\%$ positive), mainly on fine-grained tasks, where the rate rises to $83\%$. Thus, additional shots help convolutional models benefit from MC-RFM primarily when the task contains fine-grained structure.

\begin{table}[h]
\centering
\caption{Main 4-shot results on representative frozen backbones. We report Top-1 accuracy in percentage, mean $\pm$ std over three seeds. $\Delta$ is computed against the best single-geometry baseline among Euclidean
and Hyperbolic-only. Full results over all backbones and shots in Appendix~\ref{app:complete_results}.}
\label{tab:main_4shot_results}

\setlength{\tabcolsep}{4pt}
\begin{tabular}{llcccc}
\toprule
Dataset & Backbone & Euclidean & Hyperbolic-only & MC-RFM & $\Delta$ \\
\midrule
CIFAR-10 & ResNet-50 
& $63.19{\pm}1.12$ & $63.88{\pm}0.99$ & $\mathbf{64.19{\pm}1.27}$ & $+0.31$ \\
CIFAR-10 & ViT-B/16 
& $78.20{\pm}3.16$ & $81.24{\pm}2.93$ & $\mathbf{81.28{\pm}3.11}$ & $+0.04$ \\

CIFAR-100 & ResNet-50 
& $\mathbf{38.51{\pm}1.21}$ & $35.68{\pm}0.68$ & $38.47{\pm}1.13$ & $-0.04$ \\
CIFAR-100 & ViT-B/16 
& $59.37{\pm}0.87$ & $57.51{\pm}1.15$ & $\mathbf{59.41{\pm}1.05}$ & $+0.04$ \\

DTD & ResNet-50 
& $48.95{\pm}0.67$ & $48.00{\pm}0.30$ & $\mathbf{49.22{\pm}0.27}$ & $+0.27$ \\
DTD & ViT-B/16 
& $48.44{\pm}0.62$ & $48.05{\pm}0.50$ & $\mathbf{48.71{\pm}0.51}$ & $+0.27$ \\

EuroSAT & ResNet-50 
& $73.14{\pm}1.15$ & $73.25{\pm}1.58$ & $\mathbf{73.29{\pm}1.70}$ & $+0.04$ \\
EuroSAT & ViT-B/16 
& $61.33{\pm}3.79$ & $62.06{\pm}3.42$ & $\mathbf{63.76{\pm}4.68}$ & $+1.70$ \\

FGVC Aircraft & ResNet-50 
& $\mathbf{15.08{\pm}0.62}$ & $13.21{\pm}0.72$ & $14.95{\pm}0.66$ & $-0.13$ \\
FGVC Aircraft & ViT-B/16 
& $12.92{\pm}1.04$ & $11.06{\pm}0.57$ & $\mathbf{13.15{\pm}1.02}$ & $+0.23$ \\

Flowers102 & ResNet-50 
& $\mathbf{67.96{\pm}1.89}$ & $64.60{\pm}1.33$ & $67.27{\pm}1.73$ & $-0.69$ \\
Flowers102 & ViT-B/16 
& $95.52{\pm}0.33$ & $94.20{\pm}0.42$ & $\mathbf{95.63{\pm}0.35}$ & $+0.11$ \\

Food-101 & ResNet-50 
& $31.18{\pm}0.49$ & $26.00{\pm}0.63$ & $\mathbf{32.77{\pm}0.49}$ & $+1.59$ \\
Food-101 & ViT-B/16 
& $44.97{\pm}1.93$ & $40.40{\pm}1.59$ & $\mathbf{45.09{\pm}1.74}$ & $+0.12$ \\
\bottomrule
\end{tabular}
\end{table}

\subsection{Ablation analysis}
\label{sec:ablation}

We ablate the main components of MC-RFM in the 4-shot setting on five benchmarks with two frozen backbones, ResNet-50 and ViT-B/16. This gives 10 (dataset, backbone) cells per variant. Table~\ref{tab:ablation_summary} reports the mean change in Top-1 accuracy, in percentage points, relative to the full MC-RFM model.

\begin{table}[h]
\centering
\caption{Component ablation summary for MC-RFM in the 4-shot setting. Each row removes or replaces one component; values are mean Top-1 changes in percentage points relative to the full model. RN50 and ViT average over five datasets. $\#\downarrow$ reports the number of cells where the variant is strictly worse than MC-RFM. Rows are sorted by overall impact. }
\label{tab:ablation_summary}

\setlength{\tabcolsep}{4pt}
\begin{tabular}{lccc ccccc}
\toprule
& \multicolumn{3}{c}{Backbone / robustness} 
& \multicolumn{5}{c}{Per-dataset $\bar{\Delta}$ Top-1 (pp)} \\
\cmidrule(lr){2-4} \cmidrule(lr){5-9}
Component removed / replaced 
& RN50 & ViT & $\#\downarrow$ 
& FGVC & C100 & DTD & ES & F101 \\
\midrule
No cross-entropy loss                   
& $-6.09$ & $-5.16$ & 10/10 
& $-4.52$ & $-7.46$ & $-4.38$ & $-5.07$ & $-6.71$ \\

MC head $\rightarrow$ linear head    
& $-2.51$ & $-2.38$ & 10/10 
& $-2.04$ & $-1.92$ & $-1.72$ & $-1.59$ & $-4.99$ \\

MC head $\rightarrow$ prototype head 
& $-1.71$ & $-1.37$ & 10/10 
& $-0.34$ & $-1.24$ & $-1.62$ & $-1.71$ & $-2.80$ \\

No feature-adaptive branch gate         
& $-1.27$ & $-1.06$ & 10/10 
& $-0.17$ & $-0.85$ & $-0.93$ & $-1.46$ & $-2.42$ \\

No feature-adaptive $\beta$             
& $-1.31$ & $-1.01$ & 9/10 
& $-0.34$ & $-0.86$ & $-0.94$ & $-1.15$ & $-2.51$ \\

No prototype shrinkage                  
& $-1.26$ & $-0.80$ & 9/10 
& $-0.20$ & $-0.77$ & $-0.78$ & $-0.92$ & $-2.49$ \\

No task context                         
& $-0.79$ & $-1.04$ & 10/10 
& $-0.21$ & $-0.21$ & $-1.22$ & $-1.22$ & $-1.73$ \\
\bottomrule
\end{tabular}
\end{table}

\textbf{Findings.} 
The cross-entropy term is the most critical component: removing it drops performance by $-5.6$ pp on average, indicating that flow matching alone is insufficient for few-shot discrimination. The mixed-curvature head is also important, with drops of $-2.4$ pp and $-1.5$ pp when replaced by linear-only or prototype-only heads, respectively. Finally, adaptive gating, feature-adaptive $\beta$, prototype shrinkage, and task context each consistently improve performance, with removals degrading at least 9/10 cells. Overall, MC-RFM's gains arise from the combination of mixed geometry, adaptive routing, and discriminative supervision.



\textbf{Sensitivity and stability analysis.}

\begin{figure}[h]
    \centering
    \includegraphics[width=0.65\textwidth]{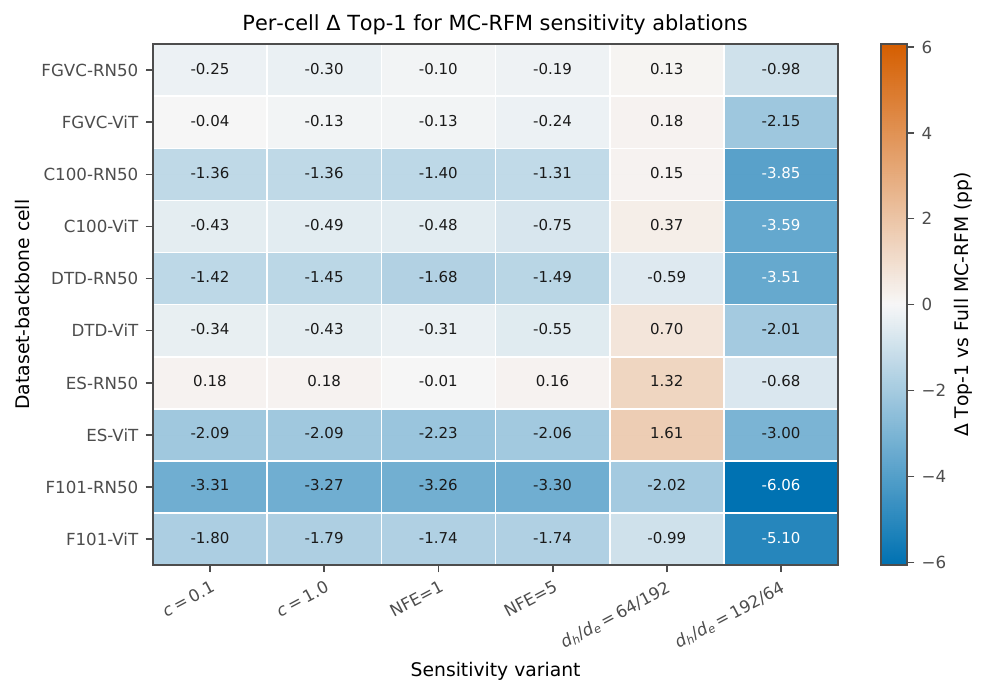}
    \caption{Sensitivity ablations of MC-RFM in the 4-shot setting. Each cell reports the Top-1 change (in percentage points) relative to the Full MC-RFM model for the same dataset and backbone. Negative values indicate performance degradation.
    MC-RFM is relatively stable to the curvature and number of ODE function evaluations, whereas allocating too much capacity to the hyperbolic branch ($d_h/d_e=192/64$) consistently hurts performance.}
    \label{fig:sensitivity_heatmap}
\end{figure}

Figure~\ref{fig:sensitivity_heatmap} evaluates sensitivity to curvature, hyperbolic-Euclidean dimensional split, and number of ODE evaluations. \textsc{MC-RFM} is relatively stable to curvature and NFE, with most changes small in magnitude. The dimensional split has a stronger effect: allocating too much capacity to the hyperbolic branch \((d_h/d_e=192/64)\) consistently hurts performance, whereas a lighter hyperbolic allocation \((d_h/d_e=64/192)\) is often competitive and sometimes improves over the default, suggesting that a compact hyperbolic subspace is sufficient to capture hierarchical structure. Across all ablation and sensitivity runs, we observe no collapse, boundary-risk, or severe loss-imbalance events; the boundary margin remains close to one, confirming that hyperbolic states stay safely inside the Poincar\'e ball, and the median \(L_h/L_e\) ratio remains bounded, with expected shifts under larger hyperbolic allocation and removal of the CE term. Thus, the observed trends reflect architectural effects rather than numerical instability. Full stability diagnostics are reported in Appendix~\ref{app:stability_diagnostics}.

\section{Discussion} \label{sec:discussion}

\subsection{Why does it perform better on specific tasks and architectures ?}

\textsc{MC-RFM} appears particularly well suited to Transformer-based backbones because its adaptation is both \emph{geometric} and \emph{incremental}: its learned ODE vector field transports features smoothly rather than through a single discrete update, consistent with flow matching as vector-field learning along prescribed probability paths \cite{lipman2023flowmatching}. This is most useful when frozen representations already provide stable semantics and well-formed neighborhoods, allowing class regions to be realigned without relearning low-level detectors. Vision Transformers naturally satisfy this condition: ViT uses globally contextualized patch tokens \cite{dosovitskiy2021vit}, while Swin captures local detail and multi-scale structure through hierarchical shifted-window attention \cite{liu2021swin}. This explains why \textsc{MC-RFM} is expected to help fine-grained tasks: hyperbolic geometry, such as the Poincaré ball, provides an efficient inductive bias for latent hierarchies, while the Euclidean factor captures non-hierarchical variation such as lighting, background, viewpoint, and style or global-shape noise \cite{nickel2017poincare,nickel2018lorentz}. Fine-grained classes often form tight semantic clusters with parent-child relations, such as aircraft manufacturer, family, and variant, with decision boundaries depending on subtle localized cues and frequent class overlap \cite{he2022transfg}. By contrast, frozen convolutional backbones remain more tied to local filter-based features \cite{naseer2021intriguing}, and coarse-grained tasks (e.g, ``cat vs. truck'') are often separable through stable global shape/layout cues, leaving less headroom for lightweight geometry-aware transport.

\subsection{Limitations}
\label{subsec:limitations}
MC-RFM is a feature-space adapter whose benefits depend on the geometry of the frozen representation. When downstream classes are already well separated, or when the task is mostly Euclidean, simpler variants can be competitive. The method also relies on support-set prototypes, which may be noisy in extreme low-shot regimes, and introduces hyperbolic design choices such as curvature, feature scaling, and dimensional split. Nevertheless, our ablations and stability diagnostics show that these components can be controlled effectively, making mixed-curvature transport a promising direction for geometry-aware few-shot adaptation.
\section{Conclusion} \label{sec:conclusion}
We introduced \textsc{MC-RFM}, a mixed-curvature Riemannian flow-matching framework for few-shot adaptation of frozen visual backbones. Rather than treating adaptation as a static Euclidean perturbation, MC-RFM models feature adaptation as task-conditioned continuous transport on a product manifold that combines hyperbolic and Euclidean structure. Across diverse datasets, backbones, and shot regimes, our results show that this geometry-aware formulation is particularly effective for Transformer-based representations and fine-grained recognition tasks. Component ablations further indicate that the gains arise from the combined effect of mixed-curvature transport, task conditioning, adaptive gating, prototype shrinkage, and discriminative supervision. Overall, MC-RFM suggests that the geometry and dynamics of representation adaptation are important design axes for few-shot learning, opening a path toward more principled and structure-aware adaptation methods.

\bibliographystyle{abbrv}
\bibliography{neurips_2026}

@article{krizhevsky_learning_2009,
	title = {Learning multiple layers of features from tiny images},
	publisher = {Toronto, ON, Canada},
	author = {Krizhevsky, Alex and Hinton, Geoffrey and {others}},
	year = {2009},
}

@inproceedings{cimpoi_describing_2014,
	title = {Describing {Textures} in the {Wild}},
	booktitle = {Proceedings of the {IEEE} {Conference} on {Computer} {Vision} and {Pattern} {Recognition} ({CVPR})},
	author = {Cimpoi, Mircea and Maji, Subhransu and Kokkinos, Iasonas and Mohamed, Sammy and Vedaldi, Andrea},
	month = jun,
	year = {2014},
}

@article{helber_eurosat_2019,
	title = {Eurosat: {A} novel dataset and deep learning benchmark for land use and land cover classification},
	volume = {12},
	number = {7},
	journal = {IEEE Journal of Selected Topics in Applied Earth Observations and Remote Sensing},
	publisher = {IEEE},
	author = {Helber, Patrick and Bischke, Benjamin and Dengel, Andreas and Borth, Damian},
	year = {2019},
	pages = {2217--2226},
}

@techreport{maji_fine-grained_2013,
	title = {Fine-{Grained} {Visual} {Classification} of {Aircraft}},
	author = {Maji, S. and Kannala, J. and Rahtu, E. and Blaschko, M. and Vedaldi, A.},
	year = {2013},
    journal = {arXiv},
	note = {\_eprint: 1306.5151},
}

@inproceedings{nilsback_automated_2008,
	title = {Automated flower classification over a large number of classes},
	booktitle = {2008 {Sixth} {Indian} conference on computer vision, graphics \& image processing},
	publisher = {IEEE},
	author = {Nilsback, Maria-Elena and Zisserman, Andrew},
	year = {2008},
	pages = {722--729},
}

@inproceedings{bossard_food-101_2014,
	address = {Cham},
	title = {Food-101 – {Mining} {Discriminative} {Components} with {Random} {Forests}},
	isbn = {978-3-319-10599-4},
	booktitle = {Computer {Vision} – {ECCV} 2014},
	publisher = {Springer International Publishing},
	author = {Bossard, Lukas and Guillaumin, Matthieu and Van Gool, Luc},
	editor = {Fleet, David and Pajdla, Tomas and Schiele, Bernt and Tuytelaars, Tinne},
	year = {2014},
	pages = {446--461},
}

@inproceedings{vinyals2016matching,
  title     = {Matching Networks for One Shot Learning},
  author    = {Vinyals, Oriol and Blundell, Charles and Lillicrap, Timothy and Kavukcuoglu, Koray and Wierstra, Daan},
  booktitle = {Advances in Neural Information Processing Systems},
  volume    = {29},
  pages     = {3630--3638},
  year      = {2016},
  url       = {https://papers.nips.cc/paper/6385-matching-networks-for-one-shot-learning}
}

@inproceedings{snell2017prototypical,
  title     = {Prototypical Networks for Few-shot Learning},
  author    = {Snell, Jake and Swersky, Kevin and Zemel, Richard S.},
  booktitle = {Advances in Neural Information Processing Systems},
  volume    = {30},
  year      = {2017},
  url       = {https://papers.nips.cc/paper/6996-prototypical-networks-for-few-shot-learning}
}

@inproceedings{finn2017maml,
  title     = {Model-Agnostic Meta-Learning for Fast Adaptation of Deep Networks},
  author    = {Finn, Chelsea and Abbeel, Pieter and Levine, Sergey},
  booktitle = {Proceedings of the 34th International Conference on Machine Learning},
  series    = {Proceedings of Machine Learning Research},
  volume    = {70},
  pages     = {1126--1135},
  publisher = {PMLR},
  year      = {2017},
  url       = {https://proceedings.mlr.press/v70/finn17a.html}
}

@inproceedings{he2016resnet,
  title     = {Deep Residual Learning for Image Recognition},
  author    = {He, Kaiming and Zhang, Xiangyu and Ren, Shaoqing and Sun, Jian},
  booktitle = {Proceedings of the IEEE Conference on Computer Vision and Pattern Recognition},
  pages     = {770--778},
  year      = {2016},
  doi       = {10.1109/CVPR.2016.90},
  url       = {https://openaccess.thecvf.com/content_cvpr_2016/html/He_Deep_Residual_Learning_CVPR_2016_paper.html}
}

@inproceedings{liu2022convnext,
  title     = {A ConvNet for the 2020s},
  author    = {Liu, Zhuang and Mao, Hanzi and Wu, Chao-Yuan and Feichtenhofer, Christoph and Darrell, Trevor and Xie, Saining},
  booktitle = {Proceedings of the IEEE/CVF Conference on Computer Vision and Pattern Recognition},
  pages     = {11966--11976},
  year      = {2022},
  doi       = {10.1109/CVPR52688.2022.01167},
  url       = {https://openaccess.thecvf.com/content/CVPR2022/html/Liu_A_ConvNet_for_the_2020s_CVPR_2022_paper.html}
}

@inproceedings{dosovitskiy2021vit,
  title     = {An Image is Worth 16x16 Words: Transformers for Image Recognition at Scale},
  author    = {Dosovitskiy, Alexey and Beyer, Lucas and Kolesnikov, Alexander and Weissenborn, Dirk and Zhai, Xiaohua and Unterthiner, Thomas and Dehghani, Mostafa and Minderer, Matthias and Heigold, Georg and Gelly, Sylvain and Uszkoreit, Jakob and Houlsby, Neil},
  booktitle = {International Conference on Learning Representations},
  year      = {2021},
  url       = {https://openreview.net/forum?id=YicbFdNTTy}
}

@inproceedings{touvron2021deit,
  title     = {Training Data-Efficient Image Transformers and Distillation through Attention},
  author    = {Touvron, Hugo and Cord, Matthieu and Douze, Matthijs and Massa, Francisco and Sablayrolles, Alexandre and J{\'e}gou, Herv{\'e}},
  booktitle = {Proceedings of the 38th International Conference on Machine Learning},
  series    = {Proceedings of Machine Learning Research},
  volume    = {139},
  pages     = {10347--10357},
  publisher = {PMLR},
  year      = {2021},
  url       = {https://proceedings.mlr.press/v139/touvron21a.html}
}

@inproceedings{liu2021swin,
  title     = {Swin Transformer: Hierarchical Vision Transformer Using Shifted Windows},
  author    = {Liu, Ze and Lin, Yutong and Cao, Yue and Hu, Han and Wei, Yixuan and Zhang, Zheng and Lin, Stephen and Guo, Baining},
  booktitle = {Proceedings of the IEEE/CVF International Conference on Computer Vision},
  pages     = {9992--10002},
  year      = {2021},
  doi       = {10.1109/ICCV48922.2021.00986},
  url       = {https://openaccess.thecvf.com/content/ICCV2021/html/Liu_Swin_Transformer_Hierarchical_Vision_Transformer_Using_Shifted_Windows_ICCV_2021_paper.html}
}

@inproceedings{houlsby2019adapters,
  title     = {Parameter-Efficient Transfer Learning for NLP},
  author    = {Houlsby, Neil and Giurgiu, Andrei and Jastrzebski, Stanislaw and Morrone, Bruna and De Laroussilhe, Quentin and Gesmundo, Andrea and Attariyan, Mona and Gelly, Sylvain},
  booktitle = {Proceedings of the 36th International Conference on Machine Learning},
  series    = {Proceedings of Machine Learning Research},
  volume    = {97},
  pages     = {2790--2799},
  publisher = {PMLR},
  year      = {2019},
  url       = {https://proceedings.mlr.press/v97/houlsby19a.html}
}

@inproceedings{hu2022lora,
  title     = {LoRA: Low-Rank Adaptation of Large Language Models},
  author    = {Hu, Edward J. and Shen, Yelong and Wallis, Phillip and Allen-Zhu, Zeyuan and Li, Yuanzhi and Wang, Shean and Wang, Lu and Chen, Weizhu},
  booktitle = {International Conference on Learning Representations},
  year      = {2022},
  url       = {https://openreview.net/forum?id=nZeVKeeFYf9}
}

@inproceedings{chen2022adaptformer,
  title     = {AdaptFormer: Adapting Vision Transformers for Scalable Visual Recognition},
  author    = {Chen, Shoufa and Ge, Chongjian and Tong, Zhan and Wang, Jiangliu and Song, Yibing and Wang, Jue and Luo, Ping},
  booktitle = {Advances in Neural Information Processing Systems},
  volume    = {35},
  year      = {2022},
  url       = {https://openreview.net/forum?id=ATiz_CDA66}
}

@inproceedings{lipman2023flowmatching,
  title     = {Flow Matching for Generative Modeling},
  author    = {Lipman, Yaron and Chen, Ricky T. Q. and Ben-Hamu, Heli and Nickel, Maximilian and Le, Matthew},
  booktitle = {International Conference on Learning Representations},
  year      = {2023},
  url       = {https://openreview.net/forum?id=PqvMRDCJT9t}
}

@inproceedings{nickel2017poincare,
  title     = {Poincar{\'e} Embeddings for Learning Hierarchical Representations},
  author    = {Nickel, Maximilian and Kiela, Douwe},
  booktitle = {Advances in Neural Information Processing Systems},
  volume    = {30},
  year      = {2017},
  url       = {https://papers.nips.cc/paper/7213-poincare-embeddings-for-learning-hierarchical-representations}
}

@inproceedings{nickel2018lorentz,
  title     = {Learning Continuous Hierarchies in the Lorentz Model of Hyperbolic Geometry},
  author    = {Nickel, Maximilian and Kiela, Douwe},
  booktitle = {Proceedings of the 35th International Conference on Machine Learning},
  series    = {Proceedings of Machine Learning Research},
  volume    = {80},
  pages     = {3779--3788},
  publisher = {PMLR},
  year      = {2018},
  url       = {https://proceedings.mlr.press/v80/nickel18a.html}
}

@inproceedings{khrulkov2020hyperbolic,
  title     = {Hyperbolic Image Embeddings},
  author    = {Khrulkov, Valentin and Mirvakhabova, Leyla and Ustinova, Evgeniya and Oseledets, Ivan and Lempitsky, Victor},
  booktitle = {Proceedings of the IEEE/CVF Conference on Computer Vision and Pattern Recognition},
  pages     = {6418--6428},
  year      = {2020},
  url       = {https://openaccess.thecvf.com/content_CVPR_2020/html/Khrulkov_Hyperbolic_Image_Embeddings_CVPR_2020_paper.html}
}

@inproceedings{ermolov2022hyperbolicvit,
  title     = {Hyperbolic Vision Transformers: Combining Improvements in Metric Learning},
  author    = {Ermolov, Aleksandr and Mirvakhabova, Leyla and Khrulkov, Valentin and Sebe, Nicu and Oseledets, Ivan},
  booktitle = {Proceedings of the IEEE/CVF Conference on Computer Vision and Pattern Recognition},
  pages     = {7409--7419},
  year      = {2022},
  doi       = {10.1109/CVPR52688.2022.00726},
  url       = {https://openaccess.thecvf.com/content/CVPR2022/html/Ermolov_Hyperbolic_Vision_Transformers_Combining_Improvements_in_Metric_Learning_CVPR_2022_paper.html}
}

@inproceedings{gu2019mixedcurvature,
  title     = {Learning Mixed-Curvature Representations in Product Spaces},
  author    = {Gu, Albert and Sala, Frederic and Gunel, Beliz and R{\'e}, Christopher},
  booktitle = {International Conference on Learning Representations},
  year      = {2019},
  url       = {https://openreview.net/forum?id=HJxeWnCcF7}
}

@inproceedings{chen2020simclr,
  title     = {A Simple Framework for Contrastive Learning of Visual Representations},
  author    = {Chen, Ting and Kornblith, Simon and Norouzi, Mohammad and Hinton, Geoffrey},
  booktitle = {Proceedings of the 37th International Conference on Machine Learning},
  series    = {Proceedings of Machine Learning Research},
  volume    = {119},
  pages     = {1597--1607},
  publisher = {PMLR},
  year      = {2020},
  url       = {https://proceedings.mlr.press/v119/chen20j.html}
}

@inproceedings{he2020moco,
  title     = {Momentum Contrast for Unsupervised Visual Representation Learning},
  author    = {He, Kaiming and Fan, Haoqi and Wu, Yuxin and Xie, Saining and Girshick, Ross},
  booktitle = {Proceedings of the IEEE/CVF Conference on Computer Vision and Pattern Recognition},
  pages     = {9729--9738},
  year      = {2020},
  url       = {https://openaccess.thecvf.com/content_CVPR_2020/html/He_Momentum_Contrast_for_Unsupervised_Visual_Representation_Learning_CVPR_2020_paper.html}
}

@inproceedings{khosla2020supcon,
  title     = {Supervised Contrastive Learning},
  author    = {Khosla, Prannay and Teterwak, Piotr and Wang, Chen and Sarna, Aaron and Tian, Yonglong and Isola, Phillip and Maschinot, Aaron and Liu, Ce and Krishnan, Dilip},
  booktitle = {Advances in Neural Information Processing Systems},
  volume    = {33},
  pages     = {18661--18673},
  year      = {2020},
  url       = {https://proceedings.neurips.cc/paper/2020/hash/d89a66c7c80a29b1bdbab0f2a1a94af8-Abstract.html}
}

@inproceedings{radford2021clip,
  title     = {Learning Transferable Visual Models From Natural Language Supervision},
  author    = {Radford, Alec and Kim, Jong Wook and Hallacy, Chris and Ramesh, Aditya and Goh, Gabriel and Agarwal, Sandhini and Sastry, Girish and Askell, Amanda and Mishkin, Pamela and Clark, Jack and Krueger, Gretchen and Sutskever, Ilya},
  booktitle = {Proceedings of the 38th International Conference on Machine Learning},
  series    = {Proceedings of Machine Learning Research},
  volume    = {139},
  pages     = {8748--8763},
  publisher = {PMLR},
  year      = {2021},
  url       = {https://proceedings.mlr.press/v139/radford21a.html}
}

@inproceedings{ganea2018hyperbolicnn,
  title     = {Hyperbolic Neural Networks},
  author    = {Ganea, Octavian-Eugen and B{\'e}cigneul, Gary and Hofmann, Thomas},
  booktitle = {Advances in Neural Information Processing Systems},
  volume    = {31},
  pages     = {5350--5360},
  year      = {2018},
  url       = {https://proceedings.neurips.cc/paper/2018/hash/dbab2adc8f9d078009ee3fa810bea142-Abstract.html}
}

@inproceedings{chami2019hgcn,
  title     = {Hyperbolic Graph Convolutional Neural Networks},
  author    = {Chami, Ines and Ying, Rex and R{\'e}, Christopher and Leskovec, Jure},
  booktitle = {Advances in Neural Information Processing Systems},
  volume    = {32},
  pages     = {4869--4880},
  year      = {2019},
  url       = {https://proceedings.neurips.cc/paper/2019/hash/0415740eaa4d9decbc8da001d3fd805f-Abstract.html}
}

@inproceedings{liu2019hgnn,
  title     = {Hyperbolic Graph Neural Networks},
  author    = {Liu, Qi and Nickel, Maximilian and Kiela, Douwe},
  booktitle = {Advances in Neural Information Processing Systems},
  volume    = {32},
  pages     = {8228--8239},
  year      = {2019},
  url       = {https://papers.neurips.cc/paper/9033-hyperbolic-graph-neural-networks}
}

@inproceedings{liu2020hyperbolicvisual,
  title     = {Hyperbolic Visual Embedding Learning for Zero-Shot Recognition},
  author    = {Liu, Shaoteng and Chen, Jingjing and Pan, Liangming and Ngo, Chong-Wah and Chua, Tat-Seng and Jiang, Yu-Gang},
  booktitle = {Proceedings of the IEEE/CVF Conference on Computer Vision and Pattern Recognition},
  pages     = {9273--9281},
  year      = {2020},
  url       = {https://openaccess.thecvf.com/content_CVPR_2020/html/Liu_Hyperbolic_Visual_Embedding_Learning_for_Zero-Shot_Recognition_CVPR_2020_paper.html}
}

@inproceedings{atigh2022hyperbolicsegmentation,
  title     = {Hyperbolic Image Segmentation},
  author    = {Atigh, Mina Ghadimi and Schoep, Julian and Acar, Erman and van Noord, Nanne and Mettes, Pascal},
  booktitle = {Proceedings of the IEEE/CVF Conference on Computer Vision and Pattern Recognition},
  pages     = {4453--4462},
  year      = {2022},
  url       = {https://openaccess.thecvf.com/content/CVPR2022/html/Atigh_Hyperbolic_Image_Segmentation_CVPR_2022_paper.html}
}

@inproceedings{skopek2020mixedvae,
  title     = {Mixed-curvature Variational Autoencoders},
  author    = {Skopek, Ondrej and Ganea, Octavian-Eugen and B{\'e}cigneul, Gary},
  booktitle = {International Conference on Learning Representations},
  year      = {2020},
  url       = {https://openreview.net/forum?id=S1g6xeSKDS}
}

@inproceedings{saezdeocarizborde2023nlgs,
  title     = {Neural Latent Geometry Search: Product Manifold Inference via Gromov-Hausdorff-Informed Bayesian Optimization},
  author    = {S{\'a}ez de Oc{\'a}riz Borde, Haitz and Arroyo, Alvaro and Morales, Ismael and Posner, Ingmar and Dong, Xiaowen},
  booktitle = {Advances in Neural Information Processing Systems},
  volume    = {36},
  year      = {2023},
  url       = {https://proceedings.neurips.cc/paper_files/paper/2023/hash/78efbc5386c5a7c241e7fcc482d3c3dc-Abstract-Conference.html}
}

@article{naseer2021intriguing,
  title={Intriguing properties of vision transformers},
  author={Naseer, Muhammad Muzammal and Ranasinghe, Kanchana and Khan, Salman H and Hayat, Munawar and Shahbaz Khan, Fahad and Yang, Ming-Hsuan},
  journal={Advances in Neural Information Processing Systems},
  volume={34},
  pages={23296--23308},
  year={2021}
}

@inproceedings{he2022transfg,
  title={Transfg: A transformer architecture for fine-grained recognition},
  author={He, Ju and Chen, Jie-Neng and Liu, Shuai and Kortylewski, Adam and Yang, Cheng and Bai, Yutong and Wang, Changhu},
  booktitle={Proceedings of the AAAI conference on artificial intelligence},
  volume={36},
  number={1},
  pages={852--860},
  year={2022}
}

@inproceedings{liu2023rectified,
  title={Flow Straight and Fast: Learning to Generate and Transfer Data with Rectified Flow},
  author={Liu, Xingchao and Gong, Chengyue and Liu, Qiang},
  booktitle={International Conference on Learning Representations (ICLR)},
  year={2023}
}

@article{tong2024cfm,
  title={Improving and Generalizing Flow-Based Generative Models with Minibatch Optimal Transport},
  author={Tong, Alexander and Fatras, Kilian and Malkin, Nikolay and Huguet, Guillaume and Zhang, Yanlei and Rector-Brooks, Jarrid and Wolf, Guy and Bengio, Yoshua},
  journal={Transactions on Machine Learning Research (TMLR)},
  year={2024}
}

@inproceedings{chen2024riemannian,
  title={Flow Matching on General Geometries},
  author={Chen, Ricky T. Q. and Lipman, Yaron},
  booktitle={International Conference on Learning Representations (ICLR)},
  year={2024}
}

@inproceedings{song2021scorebased,
  title={Score-Based Generative Modeling through Stochastic Differential Equations},
  author={Song, Yang and Sohl-Dickstein, Jascha and Kingma, Diederik P. and Kumar, Abhishek and Ermon, Stefano and Poole, Ben},
  booktitle={International Conference on Learning Representations (ICLR)},
  year={2021}
}

@article{khazem2026topolora,
  title={TopoLoRA-SAM: Topology-Aware Parameter-Efficient Adaptation of Foundation Segmenters for Thin-Structure and Cross-Domain Binary Semantic Segmentation},
  author={Khazem, Salim},
  journal={arXiv preprint arXiv:2601.02273},
  year={2026}
}

@inproceedings{karras2022elucidating,
  title={Elucidating the Design Space of Diffusion-Based Generative Models},
  author={Karras, Tero and Aittala, Miika and Aila, Timo and Laine, Samuli},
  booktitle={Advances in Neural Information Processing Systems (NeurIPS)},
  year={2022}
}

@article{khazem2026adaptertune,
  title={AdapterTune: Zero-Initialized Low-Rank Adapters for Frozen Vision Transformers},
  author={Khazem, Salim},
  journal={arXiv preprint arXiv:2603.14706},
  year={2026}
}


\appendix

\newpage
\section{Algorithmic Details}
\label{app:algorithms}
\paragraph{Training and inference workflow.}

Algorithms~\ref{alg:mcrfm_training} and~\ref{alg:mcrfm_inference} summarize the full MC-RFM procedure used in our experiments. During training, frozen backbone features are cached once, class prototypes are recomputed from the support set, and the resulting task context conditions the mixed-curvature vector field. Each minibatch is transported from its initial representation toward its class prototype using the flow-matching objective, while the transported endpoint is also supervised by the hybrid prototype-linear classifier. At inference, the support set is used only to rebuild prototypes and the task context; each query feature is projected to the product manifold, transported with the learned low-NFE ODE solver, re-projected to the interior of the Poincar\'e ball after each step, and classified using the same hybrid head.
\begin{algorithm}[h]
\caption{Training MC-RFM on a few-shot task}
\label{alg:mcrfm_training}
\begin{algorithmic}[1]
\Require Frozen backbone \(f_\psi\), support set \(\mathcal{S}\), number of classes \(K\), curvature \(c\)
\State Cache features \(h_i=f_\psi(x_i)\) for all support examples
\For{each training epoch}
    \State Project support features into \((u_h,u_e)\)
    \State Build class prototypes \(p_k=(\exp_0^c(\tilde{\mu}_{h,k}),\tilde{\mu}_{e,k})\)
    \State Encode task context \(c_{\mathcal{S}}=E_\eta(\{p_k\}_{k=1}^{K})\)
    \For{each minibatch \(\{(h_i,y_i)\}\)}
        \State Project \(h_i\) to \(z_0^i=(z_{h,0}^i,z_{e,0}^i)\)
        \State Set target \(z_1^i=p_{y_i}\)
        \State Sample \(t_i\sim \mathcal{U}(\varepsilon,1-\varepsilon)\)
        \State Interpolate \(z_{t_i}^i=\gamma_{\mathcal{M}}(z_0^i,z_1^i;t_i)\)
        \State Compute chart-space target velocities \((u_h^\star,u_e^\star)\)
        \State Predict \(v_\theta(z_{t_i}^i,t_i,c_{\mathcal{S}})\)
        \State Compute adaptive multipliers \(m_h,m_e\)
        \State Integrate the ODE from \(z_0^i\) to \(z_T^i\) for classification
        \State Compute \(\mathcal{L}_{\mathrm{FM}}+\lambda_{\mathrm{cls}}\mathcal{L}_{\mathrm{CE}}\)
        \State Update adapter, vector field, task encoder, and classifier parameters
    \EndFor
\EndFor
\end{algorithmic}
\end{algorithm}

\begin{algorithm}[h]
\caption{MC-RFM inference}
\label{alg:mcrfm_inference}
\begin{algorithmic}[1]
\Require Frozen backbone \(f_\psi\), support set \(\mathcal{S}\), query image \(x\), trained MC-RFM parameters
\State Compute support prototypes \(p_k\) and task context \(c_{\mathcal{S}}\)
\State Extract query feature \(h=f_\psi(x)\)
\State Project \(h\) to initial state \(z_0=(z_{h,0},z_{e,0})\in\mathcal{M}\)
\State Solve \(\frac{dz}{dt}=v_\theta(z,t,c_{\mathcal{S}})\) from \(t=0\) to \(t=1\)
\State Re-project the hyperbolic component to the Poincare ball after each ODE step
\State Compute hybrid logits \(\ell(z_1)\) using prototypes and the linear head
\State Return \(\arg\max_k \ell_k(z_1)\)
\end{algorithmic}
\end{algorithm}

\newpage
\section{Complete results for all the experiments} \label{app:complete_results}

\setlength{\tabcolsep}{2.5pt}
\renewcommand{\arraystretch}{0.95}
\setlength{\LTleft}{0pt}
\setlength{\LTright}{0pt}

\begin{longtable}[c]{@{}llc@{\hspace{0.7em}}c@{\hspace{0.7em}}c@{\hspace{0.7em}}c@{}}

\caption{Complete few-shot adaptation results. We report Top-1 accuracy
(mean $\pm$ std over three seeds). Best method per row is shown in bold.
Abbreviations: S = shots, FGVC = FGVC Aircraft, cn = ConvNeXt.}
\label{app:full_results} \\
\toprule
Dataset & Backbone & S & Euclidean & Hyperbolic & \textsc{MC-RFM} \\
\midrule
\endfirsthead

\toprule
Dataset & Backbone & S & Euclidean & Hyperbolic & \textsc{MC-RFM} \\
\midrule
\endhead

\midrule
\multicolumn{6}{r}{\footnotesize Continued on next page.} \\
\endfoot

\bottomrule
\endlastfoot

CIFAR-10 & ResNet-50 & 1  & $0.4117{\pm}0.0354$ & $0.4269{\pm}0.0275$ & $\mathbf{0.4271{\pm}0.0373}$ \\
CIFAR-10 & ResNet-50 & 4  & $0.6319{\pm}0.0112$ & $0.6388{\pm}0.0099$ & $\mathbf{0.6419{\pm}0.0127}$ \\
CIFAR-10 & ResNet-50 & 16 & $0.7601{\pm}0.0055$ & $0.7668{\pm}0.0073$ & $\mathbf{0.7775{\pm}0.0051}$ \\
CIFAR-10 & ViT-B/16 & 1  & $0.5027{\pm}0.0422$ & $0.5287{\pm}0.0388$ & $\mathbf{0.5291{\pm}0.0438}$ \\
CIFAR-10 & ViT-B/16 & 4  & $0.7820{\pm}0.0316$ & $0.8124{\pm}0.0293$ & $\mathbf{0.8128{\pm}0.0311}$ \\
CIFAR-10 & ViT-B/16 & 16 & $0.8911{\pm}0.0016$ & $0.8974{\pm}0.0029$ & $\mathbf{0.9060{\pm}0.0015}$ \\
CIFAR-10 & cn\_base & 1  & $0.6199{\pm}0.0682$ & $0.6211{\pm}0.0941$ & $\mathbf{0.6226{\pm}0.0726}$ \\
CIFAR-10 & cn\_base & 4  & $0.8776{\pm}0.0193$ & $\mathbf{0.8870{\pm}0.0184}$ & $0.8780{\pm}0.0154$ \\
CIFAR-10 & cn\_base & 16 & $0.9380{\pm}0.0026$ & $\mathbf{0.9404{\pm}0.0040}$ & $0.9355{\pm}0.0025$ \\
CIFAR-10 & cn\_tiny & 1  & $0.5078{\pm}0.0480$ & $\mathbf{0.5483{\pm}0.0362}$ & $0.5176{\pm}0.0234$ \\
CIFAR-10 & cn\_tiny & 4  & $\mathbf{0.8011{\pm}0.0263}$ & $0.7996{\pm}0.0134$ & $0.7939{\pm}0.0259$ \\
CIFAR-10 & cn\_tiny & 16 & $0.8776{\pm}0.0017$ & $\mathbf{0.8809{\pm}0.0081}$ & $0.8741{\pm}0.0072$ \\
CIFAR-10 & deit\_base & 1  & $0.3877{\pm}0.0665$ & $\mathbf{0.4073{\pm}0.0579}$ & $0.3851{\pm}0.0675$ \\
CIFAR-10 & deit\_base & 4  & $\mathbf{0.7104{\pm}0.0041}$ & $0.7026{\pm}0.0075$ & $0.6999{\pm}0.0021$ \\
CIFAR-10 & deit\_base & 16 & $0.8327{\pm}0.0111$ & $\mathbf{0.8404{\pm}0.0047}$ & $0.8281{\pm}0.0058$ \\
CIFAR-10 & swin\_tiny & 1  & $0.1400{\pm}0.0102$ & $0.1358{\pm}0.0131$ & $\mathbf{0.1526{\pm}0.0174}$ \\
CIFAR-10 & swin\_tiny & 4  & $0.1663{\pm}0.0191$ & $0.1434{\pm}0.0137$ & $\mathbf{0.1802{\pm}0.0264}$ \\
CIFAR-10 & swin\_tiny & 16 & $\mathbf{0.1872{\pm}0.0210}$ & $0.1293{\pm}0.0180$ & $0.1860{\pm}0.0170$ \\

\midrule
CIFAR-100 & ResNet-50 & 1  & $\mathbf{0.1898{\pm}0.0073}$ & $0.1769{\pm}0.0079$ & $0.1895{\pm}0.0074$ \\
CIFAR-100 & ResNet-50 & 4  & $\mathbf{0.3851{\pm}0.0121}$ & $0.3568{\pm}0.0068$ & $0.3847{\pm}0.0113$ \\
CIFAR-100 & ResNet-50 & 16 & $0.5455{\pm}0.0072$ & $0.5197{\pm}0.0055$ & $\mathbf{0.5458{\pm}0.0091}$ \\
CIFAR-100 & ViT-B/16 & 1  & $0.3299{\pm}0.0169$ & $0.3161{\pm}0.0172$ & $\mathbf{0.3300{\pm}0.0178}$ \\
CIFAR-100 & ViT-B/16 & 4  & $0.5937{\pm}0.0087$ & $0.5751{\pm}0.0115$ & $\mathbf{0.5941{\pm}0.0105}$ \\
CIFAR-100 & ViT-B/16 & 16 & $0.7127{\pm}0.0065$ & $0.7214{\pm}0.0032$ & $\mathbf{0.7327{\pm}0.0040}$ \\
CIFAR-100 & cn\_base & 1  & $0.4100{\pm}0.0207$ & $0.4029{\pm}0.0105$ & $\mathbf{0.4146{\pm}0.0196}$ \\
CIFAR-100 & cn\_base & 4  & $\mathbf{0.6786{\pm}0.0088}$ & $0.6723{\pm}0.0121$ & $0.6734{\pm}0.0084$ \\
CIFAR-100 & cn\_base & 16 & $0.7678{\pm}0.0011$ & $\mathbf{0.7701{\pm}0.0044}$ & $0.7679{\pm}0.0027$ \\
CIFAR-100 & cn\_tiny & 1  & $0.3144{\pm}0.0142$ & $0.2966{\pm}0.0133$ & $\mathbf{0.3149{\pm}0.0168}$ \\
CIFAR-100 & cn\_tiny & 4  & $\mathbf{0.5594{\pm}0.0015}$ & $0.5521{\pm}0.0050$ & $0.5580{\pm}0.0018$ \\
CIFAR-100 & cn\_tiny & 16 & $0.6870{\pm}0.0104$ & $\mathbf{0.6881{\pm}0.0046}$ & $0.6853{\pm}0.0094$ \\
CIFAR-100 & deit\_base & 1  & $0.2557{\pm}0.0090$ & $0.2265{\pm}0.0065$ & $\mathbf{0.2591{\pm}0.0069}$ \\
CIFAR-100 & deit\_base & 4  & $\mathbf{0.4655{\pm}0.0055}$ & $0.4471{\pm}0.0027$ & $0.4568{\pm}0.0044$ \\
CIFAR-100 & deit\_base & 16 & $\mathbf{0.6247{\pm}0.0043}$ & $0.6244{\pm}0.0026$ & $0.6162{\pm}0.0044$ \\
CIFAR-100 & swin\_tiny & 1  & $0.0207{\pm}0.0015$ & $0.0120{\pm}0.0023$ & $\mathbf{0.0212{\pm}0.0022}$ \\
CIFAR-100 & swin\_tiny & 4  & $0.0252{\pm}0.0022$ & $0.0129{\pm}0.0012$ & $\mathbf{0.0253{\pm}0.0038}$ \\
CIFAR-100 & swin\_tiny & 16 & $0.0324{\pm}0.0047$ & $0.0143{\pm}0.0009$ & $\mathbf{0.0332{\pm}0.0022}$ \\

\midrule
DTD & ResNet-50 & 1  & $0.2975{\pm}0.0192$ & $0.2998{\pm}0.0240$ & $\mathbf{0.3020{\pm}0.0199}$ \\
DTD & ResNet-50 & 4  & $0.4895{\pm}0.0067$ & $0.4800{\pm}0.0030$ & $\mathbf{0.4922{\pm}0.0027}$ \\
DTD & ResNet-50 & 16 & $0.6176{\pm}0.0032$ & $0.6131{\pm}0.0067$ & $\mathbf{0.6204{\pm}0.0064}$ \\
DTD & ViT-B/16 & 1  & $0.3076{\pm}0.0151$ & $0.2927{\pm}0.0172$ & $\mathbf{0.3080{\pm}0.0148}$ \\
DTD & ViT-B/16 & 4  & $0.4844{\pm}0.0062$ & $0.4805{\pm}0.0050$ & $\mathbf{0.4871{\pm}0.0051}$ \\
DTD & ViT-B/16 & 16 & $0.6278{\pm}0.0123$ & $0.6287{\pm}0.0136$ & $\mathbf{0.6385{\pm}0.0141}$ \\
DTD & cn\_base & 1  & $\mathbf{0.3340{\pm}0.0241}$ & $0.3335{\pm}0.0195$ & $0.3282{\pm}0.0185$ \\
DTD & cn\_base & 4  & $0.5261{\pm}0.0117$ & $\mathbf{0.5411{\pm}0.0081}$ & $0.5229{\pm}0.0035$ \\
DTD & cn\_base & 16 & $0.6738{\pm}0.0214$ & $\mathbf{0.6936{\pm}0.0134}$ & $0.6833{\pm}0.0179$ \\
DTD & cn\_tiny & 1  & $0.2842{\pm}0.0034$ & $0.2848{\pm}0.0159$ & $\mathbf{0.2882{\pm}0.0108}$ \\
DTD & cn\_tiny & 4  & $0.4839{\pm}0.0081$ & $\mathbf{0.4959{\pm}0.0034}$ & $0.4791{\pm}0.0044$ \\
DTD & cn\_tiny & 16 & $0.6376{\pm}0.0132$ & $\mathbf{0.6530{\pm}0.0130}$ & $0.6399{\pm}0.0178$ \\
DTD & deit\_base & 1  & $0.2732{\pm}0.0183$ & $0.2684{\pm}0.0317$ & $\mathbf{0.2775{\pm}0.0178}$ \\
DTD & deit\_base & 4  & $0.4482{\pm}0.0149$ & $0.4477{\pm}0.0063$ & $\mathbf{0.4486{\pm}0.0119}$ \\
DTD & deit\_base & 16 & $0.5887{\pm}0.0041$ & $0.6092{\pm}0.0053$ & $\mathbf{0.6108{\pm}0.0054}$ \\
DTD & swin\_tiny & 1  & $\mathbf{0.0420{\pm}0.0104}$ & $0.0326{\pm}0.0074$ & $0.0406{\pm}0.0059$ \\
DTD & swin\_tiny & 4  & $\mathbf{0.0495{\pm}0.0040}$ & $0.0296{\pm}0.0055$ & $0.0473{\pm}0.0087$ \\
DTD & swin\_tiny & 16 & $0.0530{\pm}0.0035$ & $0.0369{\pm}0.0016$ & $\mathbf{0.0576{\pm}0.0034}$ \\

\midrule
EuroSAT & ResNet-50 & 1  & $0.5331{\pm}0.0240$ & $0.5112{\pm}0.0385$ & $\mathbf{0.5347{\pm}0.0281}$ \\
EuroSAT & ResNet-50 & 4  & $0.7314{\pm}0.0115$ & $0.7325{\pm}0.0158$ & $\mathbf{0.7329{\pm}0.0170}$ \\
EuroSAT & ResNet-50 & 16 & $0.8413{\pm}0.0020$ & $0.8441{\pm}0.0137$ & $\mathbf{0.8455{\pm}0.0119}$ \\
EuroSAT & ViT-B/16 & 1  & $0.4074{\pm}0.0591$ & $0.4370{\pm}0.0834$ & $\mathbf{0.4411{\pm}0.0579}$ \\
EuroSAT & ViT-B/16 & 4  & $0.6133{\pm}0.0379$ & $0.6206{\pm}0.0342$ & $\mathbf{0.6376{\pm}0.0468}$ \\
EuroSAT & ViT-B/16 & 16 & $0.8383{\pm}0.0098$ & $0.8415{\pm}0.0045$ & $\mathbf{0.8476{\pm}0.0104}$ \\
EuroSAT & cn\_base & 1  & $0.4965{\pm}0.0511$ & $\mathbf{0.5014{\pm}0.0460}$ & $0.4815{\pm}0.0309$ \\
EuroSAT & cn\_base & 4  & $0.7109{\pm}0.0238$ & $0.7094{\pm}0.0344$ & $\mathbf{0.7116{\pm}0.0255}$ \\
EuroSAT & cn\_base & 16 & $0.8322{\pm}0.0063$ & $\mathbf{0.8469{\pm}0.0090}$ & $0.8387{\pm}0.0048$ \\
EuroSAT & cn\_tiny & 1  & $0.4272{\pm}0.0345$ & $\mathbf{0.4334{\pm}0.0574}$ & $0.4234{\pm}0.0487$ \\
EuroSAT & cn\_tiny & 4  & $0.6578{\pm}0.0135$ & $\mathbf{0.6621{\pm}0.0067}$ & $0.6566{\pm}0.0142$ \\
EuroSAT & cn\_tiny & 16 & $0.7974{\pm}0.0076$ & $\mathbf{0.7985{\pm}0.0112}$ & $0.7956{\pm}0.0085$ \\
EuroSAT & deit\_base & 1  & $\mathbf{0.5192{\pm}0.0300}$ & $0.4779{\pm}0.0153$ & $0.5187{\pm}0.0355$ \\
EuroSAT & deit\_base & 4  & $0.7042{\pm}0.0051$ & $0.7000{\pm}0.0123$ & $\mathbf{0.7045{\pm}0.0040}$ \\
EuroSAT & deit\_base & 16 & $\mathbf{0.8586{\pm}0.0071}$ & $0.8545{\pm}0.0117$ & $0.8500{\pm}0.0100$ \\
EuroSAT & swin\_tiny & 1  & $0.2105{\pm}0.0275$ & $0.2073{\pm}0.0203$ & $\mathbf{0.2232{\pm}0.0110}$ \\
EuroSAT & swin\_tiny & 4  & $\mathbf{0.2243{\pm}0.0163}$ & $0.1922{\pm}0.0220$ & $0.2180{\pm}0.0195$ \\
EuroSAT & swin\_tiny & 16 & $0.2277{\pm}0.0330$ & $0.1993{\pm}0.0028$ & $\mathbf{0.2472{\pm}0.0147}$ \\

\midrule
FGVC & ResNet-50 & 1  & $\mathbf{0.0788{\pm}0.0027}$ & $0.0742{\pm}0.0029$ & $0.0772{\pm}0.0041$ \\
FGVC & ResNet-50 & 4  & $\mathbf{0.1508{\pm}0.0062}$ & $0.1321{\pm}0.0072$ & $0.1495{\pm}0.0066$ \\
FGVC & ResNet-50 & 16 & $0.2549{\pm}0.0084$ & $0.2437{\pm}0.0119$ & $\mathbf{0.2605{\pm}0.0104}$ \\
FGVC & ViT-B/16 & 1  & $0.0593{\pm}0.0034$ & $0.0578{\pm}0.0022$ & $\mathbf{0.0603{\pm}0.0045}$ \\
FGVC & ViT-B/16 & 4  & $0.1292{\pm}0.0104$ & $0.1106{\pm}0.0057$ & $\mathbf{0.1315{\pm}0.0102}$ \\
FGVC & ViT-B/16 & 16 & $0.2393{\pm}0.0055$ & $0.2244{\pm}0.0107$ & $\mathbf{0.2396{\pm}0.0071}$ \\
FGVC & cn\_base & 1  & $\mathbf{0.1163{\pm}0.0051}$ & $0.1079{\pm}0.0091$ & $0.1159{\pm}0.0093$ \\
FGVC & cn\_base & 4  & $\mathbf{0.2340{\pm}0.0092}$ & $0.2139{\pm}0.0059$ & $0.2333{\pm}0.0093$ \\
FGVC & cn\_base & 16 & $0.3957{\pm}0.0057$ & $0.3953{\pm}0.0061$ & $\mathbf{0.3982{\pm}0.0088}$ \\
FGVC & cn\_tiny & 1  & $\mathbf{0.1029{\pm}0.0088}$ & $0.0964{\pm}0.0041$ & $0.1011{\pm}0.0114$ \\
FGVC & cn\_tiny & 4  & $\mathbf{0.1948{\pm}0.0074}$ & $0.1776{\pm}0.0010$ & $0.1911{\pm}0.0125$ \\
FGVC & cn\_tiny & 16 & $0.3521{\pm}0.0055$ & $0.3495{\pm}0.0062$ & $\mathbf{0.3535{\pm}0.0097}$ \\
FGVC & deit\_base & 1  & $\mathbf{0.0691{\pm}0.0049}$ & $0.0642{\pm}0.0044$ & $0.0672{\pm}0.0040$ \\
FGVC & deit\_base & 4  & $\mathbf{0.1372{\pm}0.0077}$ & $0.1197{\pm}0.0054$ & $0.1322{\pm}0.0059$ \\
FGVC & deit\_base & 16 & $0.2709{\pm}0.0064$ & $0.2497{\pm}0.0038$ & $\mathbf{0.2721{\pm}0.0073}$ \\
FGVC & swin\_tiny & 1  & $0.0145{\pm}0.0026$ & $0.0120{\pm}0.0018$ & $\mathbf{0.0151{\pm}0.0014}$ \\
FGVC & swin\_tiny & 4  & $0.0172{\pm}0.0029$ & $0.0125{\pm}0.0045$ & $\mathbf{0.0198{\pm}0.0020}$ \\
FGVC & swin\_tiny & 16 & $0.0204{\pm}0.0013$ & $0.0141{\pm}0.0005$ & $\mathbf{0.0246{\pm}0.0008}$ \\

\midrule
Flowers102 & ResNet-50 & 1  & $\mathbf{0.3824{\pm}0.0172}$ & $0.3354{\pm}0.0331$ & $0.3791{\pm}0.0206$ \\
Flowers102 & ResNet-50 & 4  & $\mathbf{0.6796{\pm}0.0189}$ & $0.6460{\pm}0.0133$ & $0.6727{\pm}0.0173$ \\
Flowers102 & ResNet-50 & 16 & {N/A} & {N/A} & {N/A} \\
Flowers102 & ViT-B/16 & 1  & $0.7529{\pm}0.0136$ & $0.7168{\pm}0.0191$ & $\mathbf{0.7533{\pm}0.0128}$ \\
Flowers102 & ViT-B/16 & 4  & $0.9552{\pm}0.0033$ & $0.9420{\pm}0.0042$ & $\mathbf{0.9563{\pm}0.0035}$ \\
Flowers102 & ViT-B/16 & 16 & {N/A} & {N/A} & {N/A}\\
Flowers102 & cn\_base & 1  & $\mathbf{0.9441{\pm}0.0086}$ & $0.9341{\pm}0.0058$ & $0.9412{\pm}0.0078$ \\
Flowers102 & cn\_base & 4  & $0.9879{\pm}0.0003$ & $0.9869{\pm}0.0021$ & $\mathbf{0.9900{\pm}0.0012}$ \\
Flowers102 & cn\_base & 16 & {N/A} & {N/A} & {N/A} \\
Flowers102 & cn\_tiny & 1  & $0.8485{\pm}0.0050$ & $0.8420{\pm}0.0107$ & $\mathbf{0.8563{\pm}0.0060}$ \\
Flowers102 & cn\_tiny & 4  & $0.9670{\pm}0.0039$ & $0.9647{\pm}0.0054$ & $\mathbf{0.9690{\pm}0.0035}$ \\
Flowers102 & cn\_tiny & 16 & {N/A} & {N/A} & {N/A} \\
Flowers102 & deit\_base & 1  & $0.3832{\pm}0.0032$ & $0.3611{\pm}0.0061$ & $\mathbf{0.3866{\pm}0.0021}$ \\
Flowers102 & deit\_base & 4  & $\mathbf{0.7432{\pm}0.0089}$ & $0.7050{\pm}0.0130$ & $0.7379{\pm}0.0086$ \\
Flowers102 & deit\_base & 16 & {N/A} & {N/A} & {N/A} \\
Flowers102 & swin\_tiny & 1  & $0.0263{\pm}0.0051$ & $0.0102{\pm}0.0054$ & $\mathbf{0.0283{\pm}0.0023}$ \\
Flowers102 & swin\_tiny & 4  & $0.0244{\pm}0.0008$ & $0.0186{\pm}0.0088$ & $\mathbf{0.0247{\pm}0.0036}$ \\
Flowers102 & swin\_tiny & 16 & {N/A} & {N/A} & {N/A} \\

\midrule
Food-101 & ResNet-50 & 1  & $0.1650{\pm}0.0024$ & $0.1416{\pm}0.0041$ & $\mathbf{0.1664{\pm}0.0003}$ \\
Food-101 & ResNet-50 & 4  & $0.3118{\pm}0.0049$ & $0.2600{\pm}0.0063$ & $\mathbf{0.3277{\pm}0.0049}$ \\
Food-101 & ResNet-50 & 16 & $0.4510{\pm}0.0071$ & $0.4175{\pm}0.0050$ & $\mathbf{0.4595{\pm}0.0040}$ \\
Food-101 & ViT-B/16 & 1  & $0.2272{\pm}0.0104$ & $0.2059{\pm}0.0132$ & $\mathbf{0.2282{\pm}0.0113}$ \\
Food-101 & ViT-B/16 & 4  & $0.4497{\pm}0.0193$ & $0.4040{\pm}0.0159$ & $\mathbf{0.4509{\pm}0.0174}$ \\
Food-101 & ViT-B/16 & 16 & $0.5948{\pm}0.0036$ & $\mathbf{0.5961{\pm}0.0054}$ & $\mathbf{0.5961{\pm}0.0056}$ \\
Food-101 & cn\_base & 1  & $0.4298{\pm}0.0230$ & $0.4117{\pm}0.0150$ & $\mathbf{0.4302{\pm}0.0190}$ \\
Food-101 & cn\_base & 4  & $0.6650{\pm}0.0176$ & $0.6449{\pm}0.0164$ & $\mathbf{0.6685{\pm}0.0171}$ \\
Food-101 & cn\_base & 16 & $\mathbf{0.7817{\pm}0.0027}$ & $0.7814{\pm}0.0049$ & $0.7767{\pm}0.0032$ \\
Food-101 & cn\_tiny & 1  & $\mathbf{0.3260{\pm}0.0202}$ & $0.3125{\pm}0.0172$ & $0.3257{\pm}0.0174$ \\
Food-101 & cn\_tiny & 4  & $\mathbf{0.5480{\pm}0.0064}$ & $0.5285{\pm}0.0099$ & $0.5448{\pm}0.0081$ \\
Food-101 & cn\_tiny & 16 & $0.6917{\pm}0.0052$ & $0.6996{\pm}0.0057$ & $\mathbf{0.7003{\pm}0.0067}$ \\
Food-101 & deit\_base & 1  & $0.1776{\pm}0.0116$ & $0.1500{\pm}0.0090$ & $\mathbf{0.1788{\pm}0.0111}$ \\
Food-101 & deit\_base & 4  & $\mathbf{0.3393{\pm}0.0195}$ & $0.3121{\pm}0.0183$ & $0.3331{\pm}0.0180$ \\
Food-101 & deit\_base & 16 & $0.5140{\pm}0.0050$ & $0.5052{\pm}0.0059$ & $\mathbf{0.5263{\pm}0.0053}$ \\
Food-101 & swin\_tiny & 1  & $0.0157{\pm}0.0004$ & $0.0106{\pm}0.0002$ & $\mathbf{0.0162{\pm}0.0009}$ \\
Food-101 & swin\_tiny & 4  & $0.0152{\pm}0.0021$ & $0.0103{\pm}0.0005$ & $\mathbf{0.0169{\pm}0.0014}$ \\
Food-101 & swin\_tiny & 16 & $0.0220{\pm}0.0019$ & $0.0129{\pm}0.0011$ & $\mathbf{0.0234{\pm}0.0009}$ \\

\end{longtable}

\section{Numerical Stability Diagnostics}
\label{app:stability_diagnostics}
\begin{figure}[h]
    \centering
    \includegraphics[width=\textwidth]{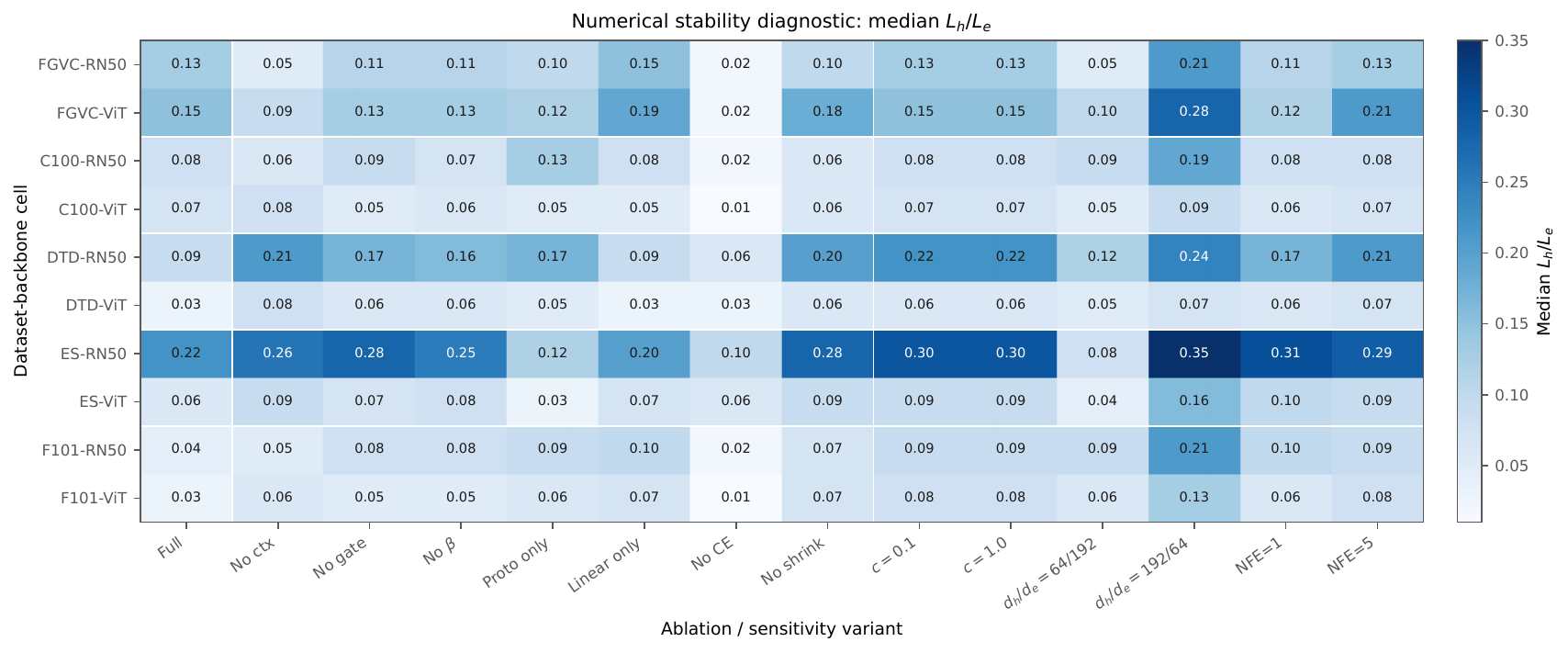}
    \caption{Numerical stability diagnostics for MC-RFM ablation and sensitivity runs. Each cell reports the median ratio between hyperbolic and Euclidean losses, \(L_h/L_e\), over three seeds. Across all runs, we observe no collapse, no boundary-risk events, and no severe loss imbalance. The bounded ratios across datasets, backbones, and ablations indicate stable mixed-curvature optimization. }
    \label{fig:stability_lh_le}
\end{figure}

We report numerical stability diagnostics for all ablation and sensitivity runs in Figure~\ref{fig:stability_lh_le}. Across all runs, MC-RFM exhibits no collapse, no boundary-risk events, and no severe loss imbalance. The boundary margin remains close to one, confirming that hyperbolic states stay safely inside the Poincar\'e ball. The median ratio \(L_h/L_e\) remains bounded across datasets, backbones, and variants, with expected increases under larger hyperbolic allocation \((d_h/d_e=192/64)\) and decreases when the CE term is removed. These diagnostics indicate that the observed ablation trends are driven by architectural choices rather than numerical instability.

\newpage
\section{Licensing and usage rights}
\label{sec:licensing}

We use publicly released pretrained weights and publicly available benchmark
datasets. Table~\ref{tab:assets_licenses} summarizes the existing assets used in
this work and the corresponding licenses or usage terms reported by the official
repositories or dataset pages. All datasets with research-only or non-commercial
usage terms are used solely for non-commercial academic evaluation.

\begin{table}[h]
\centering

\caption{Assets used in this work and their corresponding licenses.}
\label{tab:assets_licenses}
\begin{tabular}{lll}
\toprule
\textbf{Asset} & \textbf{Type} & \textbf{License} \\
\midrule
ViT-B/16 & Vision Transformer model & Apache License 2.0 \\
DeiT-base & Vision Transformer model & Apache License 2.0 \\
Swin-Tiny & Vision Transformer model & MIT License \\
ConvNeXt-Tiny & Convolutional backbone & Apache License 2.0 \\
ConvNeXt-Base & Convolutional backbone & Apache License 2.0 \\
ResNet-50 & Convolutional backbone & BSD 3-Clause License \\
\midrule
CIFAR-10 & Image classification dataset & MIT License \\
CIFAR-100 & Image classification dataset & MIT License \\
FGVC-Aircraft & Fine-grained classification dataset & Non-commercial research use \\
Flowers102 & Fine-grained classification dataset & MIT License \\
DTD & Texture classification dataset & Apache License 2.0 \\
EuroSAT & Remote sensing dataset & MIT License \\
Food-101 & Food recognition dataset & MIT License \\
\bottomrule
\end{tabular}
\end{table}

\end{document}